\documentclass[preprint]{imsart}
\pdfoutput=1

\usepackage[numbers]{natbib}
\usepackage{graphicx} 
\usepackage[justification=centering]{subfig}
\usepackage{morefloats}
\usepackage{amsmath,amssymb} 
\usepackage{siunitx}
\usepackage{multirow}
\usepackage{booktabs}
\usepackage{hyperref}
\usepackage{color}
\usepackage{array}
\usepackage{tabularx}
\usepackage{xifthen}
\newcolumntype{Y}{>{\centering\arraybackslash}X}
\usepackage{dcolumn}
\newcolumntype{d}[1]{D{.}{.}{#1}}
\newcolumntype{.}{D{.}{.}{-1}}

\usepackage[table]{xcolor}

\graphicspath{{figs/}}

\definecolor{scoreAcolor}{rgb}{1,0.9,0.9}
\definecolor{scoreBcolor}{rgb}{0.9,1,0.9}
\definecolor{scoreCcolor}{rgb}{0.9,0.9,1}
\definecolor{scoreDcolor}{rgb}{1,1,0.9}

\newcommand{\set}[1]{{\mathcal N}}

\newcommand{\F}[2]{\ensuremath{#1_{#2}}}
\newcommand{\ti}[0]{0}
\newcommand{\tc}[0]{t}
\newcommand{\tp}[0]{t-1}
\newcommand{\tn}[0]{t+1}

\newcommand{\RR}{\mathbb{R}}

\begin{document}

\begin{frontmatter}

\title{Video Human Segmentation using Fuzzy Object Models and its
  Application to Body Pose Estimation of Toddlers for Behavior
  Studies\thanksref{t2}}
\runtitle{Video Human Segmentation for Body Pose Estimation of Toddlers}

  \thankstext{t2}{Manuscript submitted to IEEE Transactions on Image
    Processing on May 4, 2013. Copyright transferred to IEEE as part
    of the submission process.}

\begin{aug}
  \author{
    \fnms{Thiago Vallin} \snm{Spina}%
    \corref{}%
    \ead[label=e2]{tvspina@ic.unicamp.br}
  }

  \address{
    Institute of Computing, University of Campinas, Brazil\\
    \printead{e2}
  }

  \author{
    \fnms{Mariano} \snm{Tepper}%
  }

  \address{
    Department of Electrical and Computer Engineering, Duke University, USA.
  }

  \author{
    \fnms{Amy} \snm{Esler}
  }

  \address{
    Department of Pediatrics, University of Minnesota, USA.
  }

  \author{
    \fnms{Vassilios} \snm{Morellas}
  }

  \author{
    \fnms{Nikolaos} \snm{Papanikolopoulos}
  }

  \address{
    Department of Computer Science and Engineering, University of Minnesota, USA.
  }

  \author{
    \fnms{Alexandre Xavier} \snm{Falc{\~{a}}o}
  }

  \address{
    Institute of Computing, University of Campinas, Brazil
  }

  \author{
    \fnms{Guillermo} \snm{Sapiro}
  }

  \address{
    Department of Electrical and Computer Engineering, Department of Computer Science, and Department of Biomedical Engineering, Duke University, USA.
  }

  \runauthor{T. V. Spina et al.}

\end{aug}


\begin{abstract}
Video object segmentation is a challenging problem due to the presence
of deformable, connected, and articulated objects, intra- and
inter-object occlusions, object motion, and poor lighting. Some of
these challenges call for object models that can locate a desired
object and separate it from its surrounding background, even when both
share similar colors and textures. In this work, we extend a fuzzy
object model, named \emph{cloud system model} (CSM), to handle video
segmentation, and evaluate it for body pose estimation of toddlers at
risk of autism. CSM has been successfully used to model the parts of
the brain (cerebrum, left and right brain hemispheres, and cerebellum)
in order to automatically locate and separate them from each other,
the connected brain stem, and the background in 3D MR-images. In our
case, the objects are articulated parts (2D projections) of the human
body, which can deform, cause self-occlusions, and move along the
video. The proposed CSM extension handles articulation by connecting
the individual clouds, body parts, of the system using a 2D stickman
model. The stickman representation naturally allows us to extract 2D
body pose measures of arm asymmetry patterns during unsupported gait
of toddlers, a possible behavioral marker of autism. The results show
that our method can provide insightful knowledge to assist the
specialist's observations during real in-clinic assessments.
\end{abstract}


\end{frontmatter}

\section{Introduction}
\label{sec:intro}

The content of a video (or image) may be expressed by the objects
displayed in it, which usually possess three-dimensional
shapes. Segmenting the 2D projections of those objects from the
background is a process that involves \emph{recognition} and
\emph{delineation}. Recognition includes approximately locating the
whereabouts of the objects in each frame and verifying if the result
of delineation constitutes the desired entities, while delineation is
a low-level operation that accounts for precisely defining the
objects' spatial extent. This image processing operation is
fundamental for many applications and constitutes a major challenge
since video objects can be deformable, connected, and/or articulated;
suffering from several adverse conditions such as the presence of
intra- and inter-object occlusions, poor illumination, and color and
texture similarities with the background. Many of these adversities
require prior knowledge models about the objects of interest to make
accurate segmentation feasible.

In interactive image and video object segmentation, for example, the
model representing where (and what) are the objects of interest comes
from the user's knowledge and input (e.g., user drawn strokes), while
the computer performs the burdensome task of precisely delineating
them~\cite{Falcao98,Bai10,Bai09b,Price09}. Cues such as optical flow,
shape, color and texture are then used to implicitly model the object
when propagating segmentation throughout consecutive frames, with the
user's knowledge remaining necessary for corrections. The same type of
cues have been used to implicitly model deformable objects in
semi-supervised object tracking~\cite{Minetto12}, to overcome
adversities such as total occlusions. Unsupervised approaches often
consider motion to do pixel-level segmentation by implictly modeling
deformable and articulated video objects as coherently moving points
and regions~\cite{Ochs11}. The simple representation of a human by a
3D articulated stickman model has been used as an explicit shape
constraint by PoseCut~\cite{Kohli08} to achieve simultaneous
segmentation and body pose estimation in video. Active Shape Models
(ASMs) consider the statistics of correspondent control points
selected on training shapes to model an object of interest, in order
to locate and delineate it in a new test
image~\cite{Cootes95,Liu09}. The well-defined shapes of objects in
medical imaging has further led to the development of \emph{fuzzy
  objects models} (FOMs) to do automatic brain image
segmentation~\cite{Miranda09b,Miranda10-TR-IC} and automatic anatomy
recognition~\cite{Udupa11,Udupa12} in static 3D scenes. FOMs are able
to separate connected objects with similar color and texture from each
other and the background, while not dealing with the control point
selection and correspondence determination required for ASMs.

In this work, we propose an extension of the Cloud System Model (CSM)
framework~\cite{Miranda10-TR-IC} to handle 2D articulated bodies for
the task of segmenting humans in video. The CSM is a fuzzy object
model that aims at acting as the human operator in segmentation, by
synergistically performing recognition and delineation to
automatically segment the objects of interest in a test image or
frame. The CSM is composed of a set of correlated \emph{object
  clouds/cloud images}, where each cloud (\emph{fuzzy object})
represents a distinct object of interest. We describe the human body
using one cloud per body part in the CSM (e.g., head, torso, left
forearm, left upper arm) --- for the remainder of the paper, we shall
refer to ``object'' as a body part constituent of the cloud system. A
cloud image captures shape variations of the corresponding object to
form an uncertainty region for its boundary, representing the area
where the object's real boundary is expected to be in a new test image
(Figure~\ref{f.pose-estimation}). Clouds can be seen as global shape
constraints that are capable of separating connected objects with
similar color and texture.
\newcommand{\figoneheight}{0.14\textheight} 
\begin{figure*}[!ht]
\centering \renewcommand{\arraystretch}{0.5}
\begin{tabular}{@{\hspace{0pt}}c@{\hspace{8pt}}c@{\hspace{4pt}}c@{\hspace{4pt}}c@{\hspace{4pt}}c@{\hspace{0pt}}}

\includegraphics[height=\figoneheight]{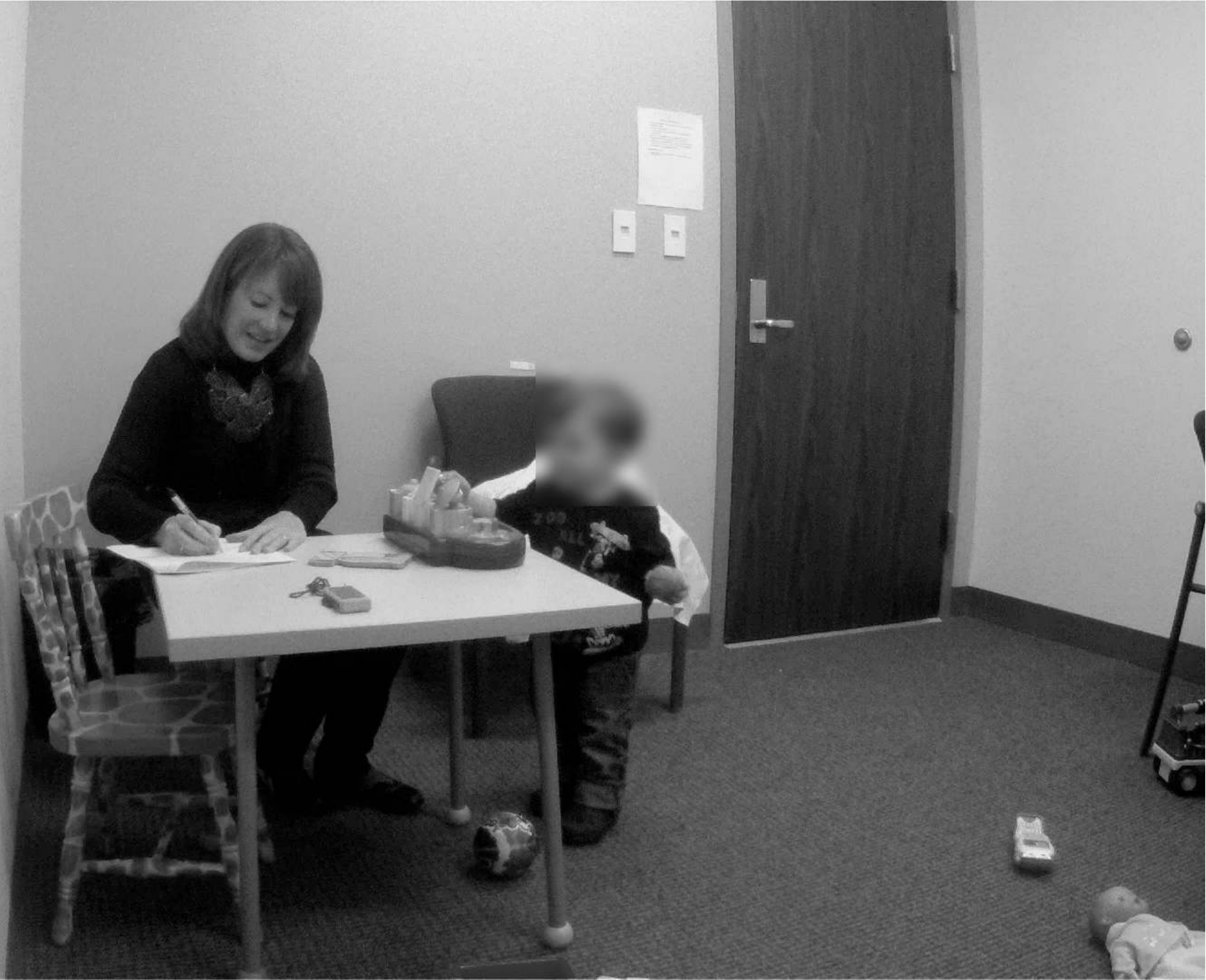} &
\subfloat[]{
    \includegraphics[height=\figoneheight]{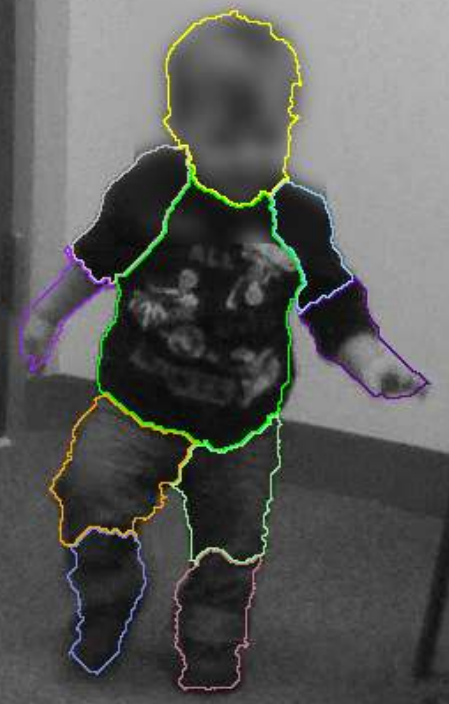}
    \label{f.pose-estimation1_seg}
} &
\subfloat[]{
    \includegraphics[height=\figoneheight]{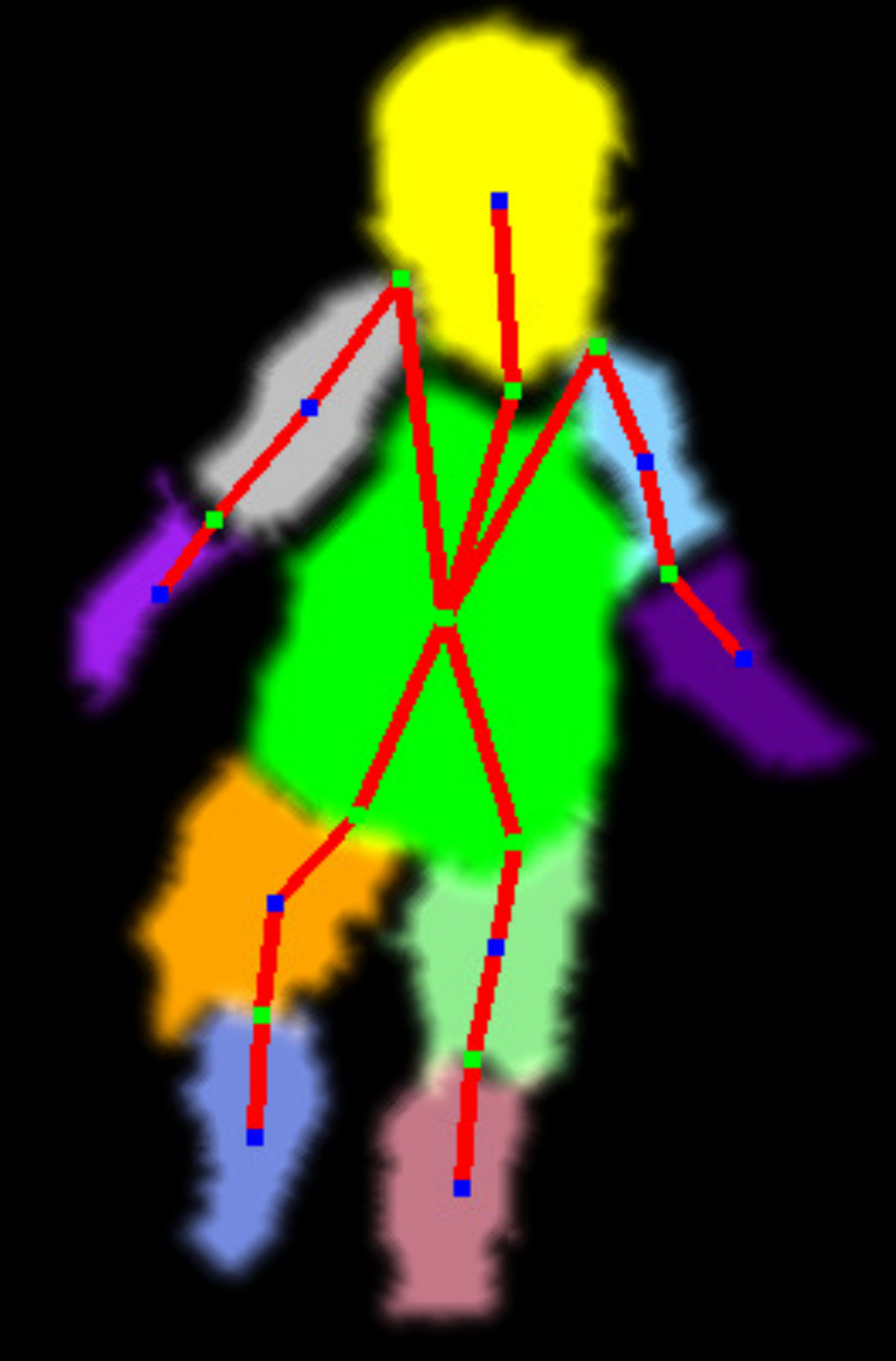}
    \label{f.pose-estimation1_stick}
} &
\subfloat[]{
    \includegraphics[height=\figoneheight]{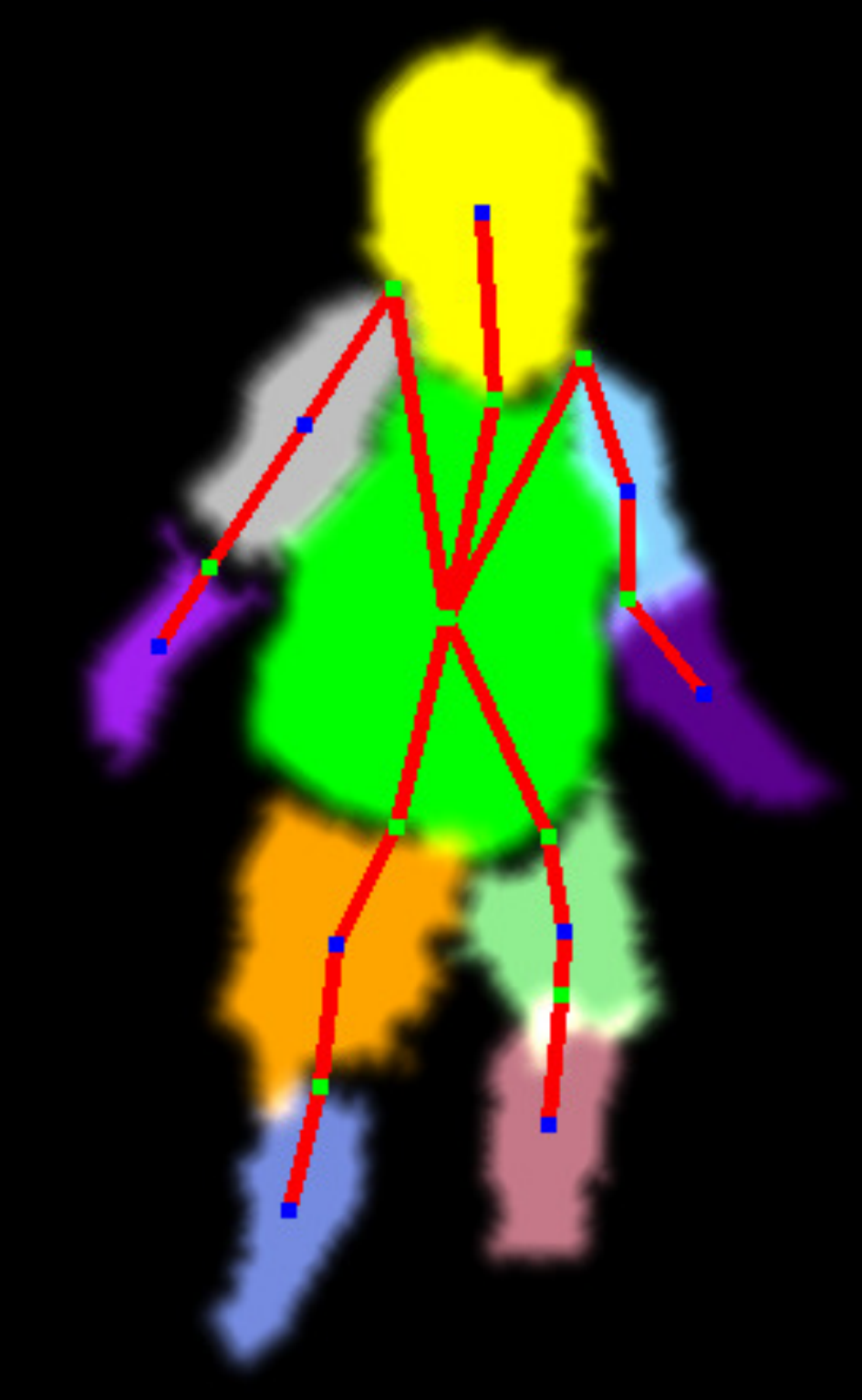}
    \label{f.pose-estimation2_seg}
} &
\subfloat[]{
    \includegraphics[height=\figoneheight]{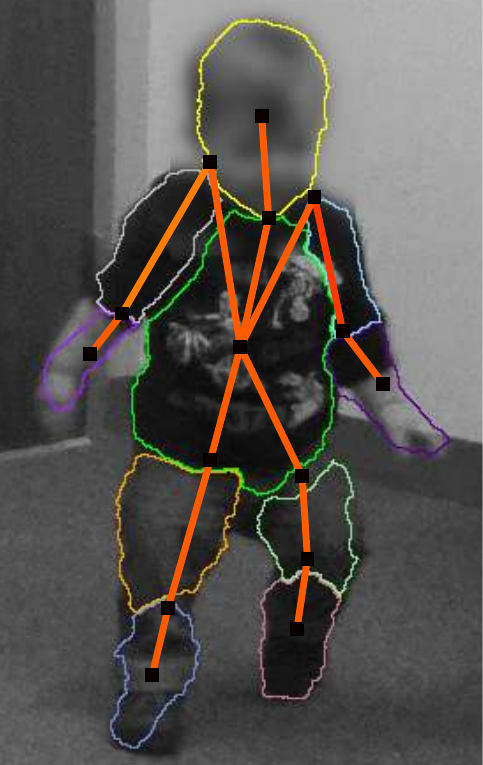}
    \label{f.pose-estimation2_stick}
}
\end{tabular}

\caption{{\bf Left:} General scene capturing free play
  activities. {\bf Right:} Overall segmentation and pose tracking
  scheme. \protect\subref{f.pose-estimation1_seg} Segmentation mask
  $\F{L}{\ti}$ provided at an initial frame
  $t=0$. \protect\subref{f.pose-estimation1_stick} CSM computed from
  $\F{L}{\ti}$ and the 2D stickman used to connect the clouds
  corresponding to each body
  part. \protect\subref{f.pose-estimation2_seg} Transformed CSM at
  frame $8$. \protect\subref{f.pose-estimation2_stick} Segmentation
  and final pose estimation. Faces blurred for privacy protection.}
\label{f.pose-estimation}
\end{figure*}

For each search position in an image, CSM executes a \emph{delineation
  algorithm} in the uncertainty regions of the clouds and evaluates if
the resulting candidate segmentation masks yield a maximum score for a
given \emph{object recognition functional}. Our recognition functional
takes into account information from previous frames and is expected to
be maximum when the uncertainty regions are properly positioned over
the real objects' boundaries in the test image (e.g.,
Figure~\ref{f.msps-search-diagram}). As originally proposed
in~\cite{Miranda10-TR-IC}, if the uncertainty regions are well adapted
to the objects' new silhouettes and the delineation is successful, the
search is reduced to translating the CSM over the image. The CSM
exploits the relative position between the objecs to achieve greater
effectiveness during the search~\cite{Miranda10-TR-IC}. Such static
approach works well for 3D brain image segmentation because the
relative position between brain parts is fairly constant and they do
not suffer from self-occlusions or foreshortening, as opposed to the
2D projections of body parts in video.

To deal with human body articulation, we extended the CSM definition
to include a hierarchical relational model in the form of a 2D
stickman rooted at the torso, that encompases how the clouds are
connected,
the relative angles between them, and their scales (similarly
to~\cite{Udupa11} and~\cite{Udupa12}). Instead of requiring a set of
label images containing delineations of the human body, as originally
needed for training fuzzy object
models~\cite{Miranda09b,Miranda10-TR-IC,Udupa11,Udupa12}, we adopt a
generative approach for human segmentation to cope with the large
variety of body shapes and poses. We create the CSM from a single
segmentation mask interactively obtained in a given initial frame
(figures~\ref{f.pose-estimation}\subref{f.pose-estimation1_seg}-\subref{f.pose-estimation1_stick}). Then,
the resulting CSM is used to automatically find the body
frame-by-frame in the video segment
(figures~\ref{f.pose-estimation}\subref{f.pose-estimation2_seg}-\subref{f.pose-estimation2_stick}). During
the search, we translate the CSM over the image while testing
different angle and scale configurations for the clouds to try a full
range of 2D body poses.

A straightforward result of using a 2D stickman to guide the CSM is
that the best configuration for segmentation directly provides a
skeleton representing the 2D body pose. Therefore, we validate our
method in the detection of early bio-markers of autism from the body
pose of at-risk toddlers~\cite{Hashemi12a,Hashemi12b}, while other
applications are possible. Motor development has often been
hypothesized as an early bio-marker of Autism Spectrum Disorder
(ASD). In particular, Esposito et al.~\cite{Esposito11} have found,
after manually performing a burdensome analysis of body poses in early
home video sequences, that toddlers diagnosed with autism often
present asymmetric arm behavior when walking unsupportedly. We aim at
providing a simple, semi-automatic, and unobtrusive tool to aid in
such type of analysis, which can be used in videos from real in-clinic
(or school) assessments for both research and diagnosis. A preliminary
version of this work partially appeared in~\cite{Hashemi12a}.
\begin{figure*}[!htb]
\begin{center}
\includegraphics[width=\textwidth]{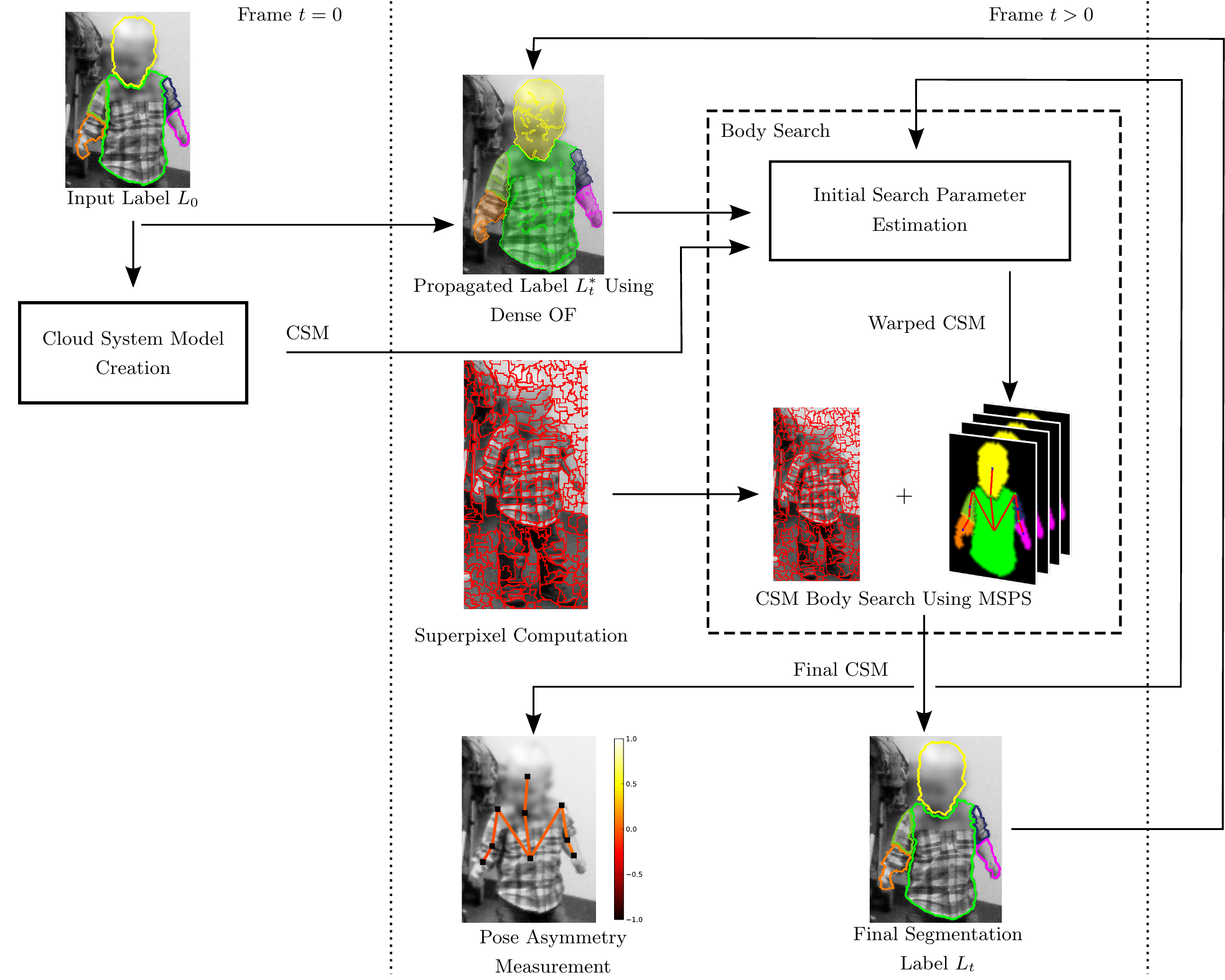} 
\end{center}
\caption{Human body segmentation and pose estimation in consecutive
  video frames using the Cloud System
  Model.\label{f.framework-diagram}}
\end{figure*}

Human body pose estimation is a complex and well explored research
topic in computer vision~\cite{Kohli08,Zuffi12,Eichner12,Ionescu11},
although it has been mostly restricted to adults, often in constrained
scenarios, and never before exploited in the application we address.
Although PoseCut~\cite{Kohli08} performs simultaneous segmentation and
body pose estimation, a key difference is that our method uses the
body shape observed from a generative mask of a single image to
concurrently track and separate similar-colored body parts
individually, whose delineation can be further evaluated for ASD risk
signs, while considering an arbitrarily complex object recognition
functional for such purpose (CSM may also consider a training dataset
of body shapes if available). Notwithstanding, our focus is to present
the extension of CSM segmentation in video, a side-effect being the
pose estimation of humans. Fuzzy object models based on the CSM could
also be used for object tracking and image-based 3D rendering, for
example. Lastly, range camera data can be easily incorporated into our
system, although this work focuses on the 2D case given the nature of
our data acquisition (the clinician repositions the camera at will to
use the videos in her assessment).

Once the skeleton (CSM stickman) is computed for each video sequence
frame, we extract simple angle measures to estimate arm asymmetry. In
this work, we treat the arm asymmetry estimation as an application for
the 2D body pose estimation, while hypothesizing that action
recognition methods based on pose and/or point trajectory
analysis~\cite{Yao12,Sivalingam12} can be further used to
automatically detect and measure other possibly stereotypical motor
behaviors (e.g., walking while holding the arms parallel to the ground
and pointing forward, arm-and-hand flapping). 

Our contributions are threefold:
\begin{enumerate}
  \item We provide an extension of the Cloud System Model to segment
    articulated bodies (humans) in video.
  \item The result of our segmentation method automatically provides
    2D body pose estimation.
  \item We validate and apply our work in the body pose estimation of
    toddlers to detect and measure early bio-markers of autism in
    videos from real in-clinic assessments.
\end{enumerate}

Section~\ref{s.articulated-csm} describes the creation of the
articulated CSM, as well as its usage for automatically segmenting the
toddler's body in a new frame. Section~\ref{s.body-pose-search}
further describes particular details regard using CSM in video to
locate and segment the human body. Finally,
Section~\ref{s.autism-assessment} explains how this work aids autism
assessment, while Section~\ref{s.results-skeleton} provides
experiments that validate our method in determining arm asymmetry.

\newcommand{\sy}{\ensuremath{s^y}}
\newcommand{\sx}{\ensuremath{s^x}}
\newcommand{\tcand}[1]{\ensuremath{\widetilde{#1}}}

\section{Articulated Cloud System Model}
\label{s.articulated-csm}

Figure~\ref{f.framework-diagram} depicts the overall scheme of our
human body segmentation method in video using the Cloud System Model.
We generate the model from a segmentation mask $\F{L}{\ti}(x)$ of the
toddler's body obtained, e.g., interactively~\cite{Spina11}, at a
given initial frame $\F{I}{\ti}(x)$ (assuming time $t=\ti$ as the
starting point). Then, in frame $\F{I}{\tc}$, $\tc>\ti$, the automatic
search for the human involves maximizing a recogntion functional by
applying affine transformations to each CSM cloud, considering the
body's tree hierarchy, until the model finds and delineates the body
in its new pose. The following subsections explain these two processes
in details.
\begin{figure*}[!htb]
\begin{center}
\includegraphics[width=\textwidth]{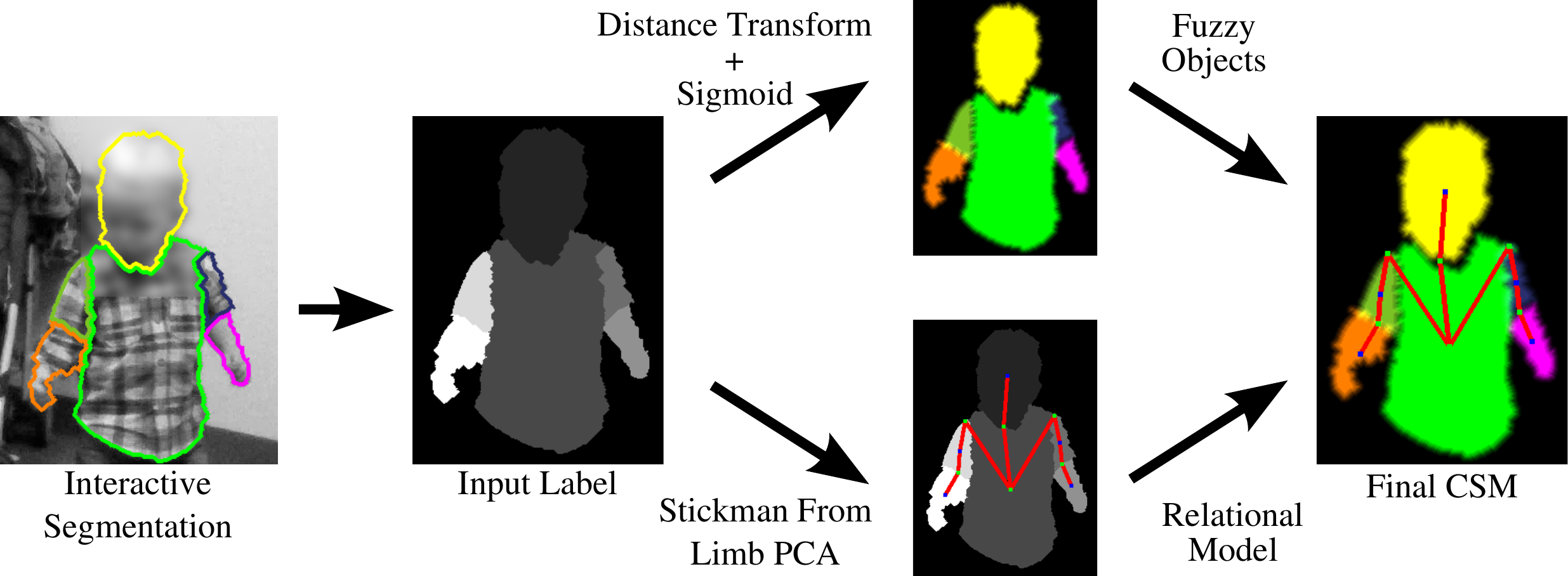} 
\end{center}
\caption{Overview of a Cloud System Model
  computation.\label{f.csm-creation-diagram}}
\end{figure*}

\subsection{Cloud System Model Creation}
\newcommand{\lb}[1]{\texttt{#1}}


Formally, the CSM is a triple $C=\{{\cal O}, A, F\}$, composed of a
set ${\cal O}$ of clouds $O_l$ (i.e., ${\cal O}$ is a \emph{cloud
  system}), a delineation algorithm $A$, and an object recognition
functional $F$~\cite{Miranda10-TR-IC}. A cloud $O_l$ is an image that
encodes the fuzzy membership $O_l(x)\in [0,1]$ that pixel $x$ has of
belonging to object $l\in\{1,\ldots,m\}$. Pixels with $O_l(x)=1$ or
$O_l(x)=0$ belong to the \emph{object} or \emph{background} regions of
the cloud, respectively, while pixels with $0<O_l(x)<1$ are within the
\emph{uncertainty region} ${\cal U}_l$. During the search, for every
location hypothesis, algorithm $A$ is executed inside the uncertainty
region ${\cal U}_l$, projected over the search frame, to extract a
candidate segmentation mask of the object $l$ from the background. We
then evaluate the set of labeled pixels ${\cal M}_l$ for all $m$ masks
using a functional $F:{\cal M}\rightarrow \RR$, and combine the
individual recognition scores $F_l$ to determine whether the body has
been properly detected/segmented. $F$ takes into account temporal
information, as will be detailed in Section~\ref{s.body-pose-search},
while Section~\ref{ss.csm-delineation} describes algorithm $A$.



We compute the cloud system ${\cal O}$ from a segmentation mask
$\F{L}{\ti}(x)\in\{0,1,\ldots,m\}$ 
where each label $l\in \F{L}{\ti}$ represents a distinct object/body
part, and $l=0$ is the background
(Figure~\ref{f.csm-creation-diagram}). Since here we are mostly
interested in the upper body to compute arm asymmetry, the body parts
represented in our CSM are: the head, torso, left and right upper
arms, and left and right forearms ($m=6$) --- with a slight abuse of
notation, we shall use $l=\lb{Torso}$ to denote the torso's label id,
for example. It should be noted, however, that our model is general
enough to segment other body parts (e.g.,
figures~\ref{f.pose-estimation} and~\ref{f.segmentation-results}),
including extremities if desired (hands and feet).

 
We first apply a signed Euclidian distance transform~\cite{Falcao02}
to the border of each body part label $l\in \F{L}{\ti}$ independently
(Figure~\ref{f.csm-creation-diagram}), generating distance maps
$DT_l(x)$ (with negative values inside the objects).
Afterwards, all distance maps $DT_l$ are smoothed to output each cloud
image $O_l\in{\cal O}$ by applying the sigmoidal function
\begin{eqnarray}
O_l(x) & = & \left\{\begin{array}{ll} 
0 & \mbox{if } DT_l(x) \geq \gamma_p, \\ 
1 & \mbox{if } DT_l(x) \leq \gamma_n, \label{e.sigmoid}\\ 
\frac{1.0}{1.0+\exp{(\frac{DT_l(x)}{\sigma_l}})} & \mbox{ otherwise,}
\end{array}\right.
\end{eqnarray}
where $\gamma_p$, $\gamma_n$, and $\sigma_l$ are parameters used to
control the size and fuzziness of the uncertainty
region.\footnote{Note that our generative approach can be readily
  complemented by having a dataset with training masks from a wide
  variety of body shapes and poses to compute the CSM. The training
  masks should represent the body of a toddler (or toddlers) with
  different poses, which would then be clustered according to the
  shapes' similarities~\cite{Miranda10-TR-IC} to yield multiple cloud
  systems ${\cal O}_g$. The cloud systems would perform the search
  simultaneously and the one with best recognition score would be
  selected~\cite{Miranda10-TR-IC}.} Typically, we define
$\sigma_l=1.5$, $\gamma_p=5$, and $\gamma_n=-4$.

\subsection{Relational Model for Articulated CSM}

\begin{figure}[htb]
\centering
\includegraphics[width=.6\textwidth]{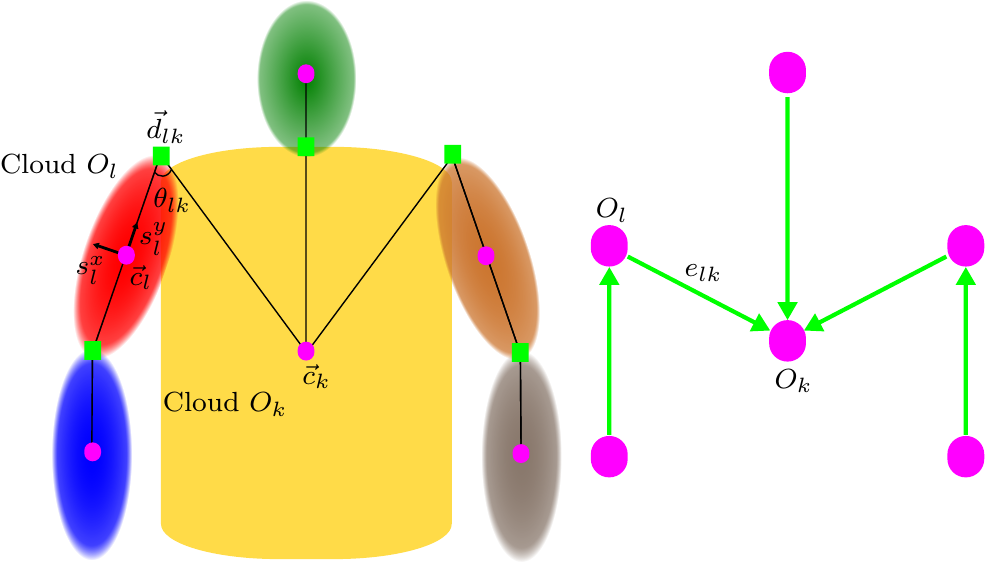} 
\caption{\textbf{Left:} The CSM representation of the upper body with
  one object cloud $O_l$ per body part. The attributes $\sx_l$ and
  $\sy_l$ are the current axis scales (w.r.t. $l$'s original size in
  frame $\F{I}{\ti}$), $\theta_{lk}$ is the current joint angle
  between $O_l$ (e.g., right upper arm) and its parent $O_k$ (torso),
  while $\vec{c}_l$, $\vec{d}_{lk}$, and $\vec{c}_k$ represent
  relative displacements among the centroid of $O_l$, the joint
  $e_{lk}$, and the centroid of $O_k$, respectively.
\textbf{Right:} The representation of the relational model $G$ in
graph notation, where the arrows indicate the predecessor relationship
between body parts.
  \label{f.relational-model}}
\end{figure}

We extend the CSM definition $C=\{{\cal O},A,F,G\}$ to include an
articulated relational model $G=\{{\cal O},{\cal E},\Omega,\Theta\}$
(attributed graph). Graph $G$ can be depicted as a 2D stickman in the
form of a tree rooted at the torso
(Figure~\ref{f.relational-model}). Each cloud in $G$ is connected to
its parent cloud/body part by the body joint between them (i.e., the
neck joint, elbow, and shoulder --- we add the hip joint, knee, wrist,
and ankle when applicable).

The nodes of $G$ are the clouds $O_l\in{\cal O}$, while the edge
$e_{lk}\in {\cal E}$ represents the body joint that connects the
clouds $O_l,O_k\in{\cal O}$ ($k$ being the predecessor of $l$ in $G$,
denoted by $P(l)=k$). $\Omega_l=(\sy_l,\sx_l,\vec{c}_l)$ defines a set
of attributes for node $l$ containing the current scales of the
primary and secondary axes of cloud $O_l$ ($\sy_l$ and $\sx_l$,
respectively), w.r.t. the original size of $O_l$ in frame
$\F{I}{\ti}$, and the cloud's centroid relative displacement
$\vec{c}_l$ to the joint $e_{lk}\in{\cal E}$, see
Figure~\ref{f.relational-model}. Similarly,
$\Theta_{lk}=(\theta_{lk},\vec{d}_{lk})$ is a set of attributes for
the body joint/edge $e_{lk}\in {\cal E}$ comprised by the relative
angle $\theta_{lk}$ between nodes $l$ and $k$, and the relative
displacement $\vec{d}_{lk}$ of the joint w.r.t. the centroid of the
predecessor cloud $O_{k}$.  We refer to $e_{lk}$ as the parent joint
of node/body part $l$. For node $l=\lb{Torso}$, we define by
convention $e_{lk}=e_{ll}\in {\cal E}$, $P(l)=nil$, $\theta_{lk}$ as
the cloud's global orientation, and $\vec{c_l}=\vec{d}_{lk}$ as the
current search position in image coordinates. The relative
displacements, scales, and angles are used to reposition the clouds
during the body search in a new frame.

The initialization of $G$ in frame $\F{I}{\ti}$ requires to determine
a suitable position for each body joint. One may simply compute the
parent joint of body part $l$ by considering it to be on the primary
axis of cloud $O_l$, in the intersection between the uncertainty
regions of $O_l$ and $O_k$ and simultaneously closer to the centroids
$c_l$ and $c_k$ of both clouds. For such purpose, we assume that the
global orientation of $O_l$ is the same of body part $l$ in the
coordinates of image $\F{L}{\ti}$, and compute it using Principal
Component Analysis (PCA) of all pixel coordinates $x$ such that
$\F{L}{\ti}(x)=l$. Such an approach implicitely assumes that the body
parts are approximately ``rectangular.'' This assumption works well
for the head and torso, thus allowing us to compute the neck joint.

Since the limb proportions of toddlers are different than those of the
adults, the aforementioned assumption is often violated, forcing us to
take a different approach to compute the orientations of each limb's
body parts. We denote $O_{i,l}$ the cloud $O_l\in {\cal O}$
representing body part $l$ from limb $i$ (e.g., limb $i=1$ contains
all the clouds for the left arm: $\lb{Left Upper Arm}$ and $\lb{Left
  Forearm}$ --- again, $\lb{Left Hand}$ can be part of the left
arm). We binarize $\F{L}{\ti}$ considering all the labels from limb
$i$ to output a mask for every limb in the CSM. Afterwards, we compute
the morphological skeleton of the binary mask~\cite{Falcao02} and use
the skeleton pixels that intersect each cloud $O_{i,l}$ of $i$ to
determine the corresponding global orientations using PCA.

The rationale behind only considering the morphological skeleton
pixels is that the skeleton closely follows the body parts' primary
axes.
The relative displacement vectors $\vec{c}_l$ and $\vec{d}_{lk}$ and
relative angle $\theta_{lk}$ can be straightforwardly computed for $G$
once we have all body joints, cloud orientations and centroid pixels
in the coordinate system of $\F{L}{\ti}$.

\subsection{Automatically Searching for the Human Body in a Frame}
\label{ss.msps-search}

Let $I$ be an image where the toddler's body is supposed to be
segmented and searched. Automatically finding the human body in $I$
using the CSM $C$, corresponds to determining the optimal state of
graph $G$ that reconfigures the clouds of $C$ in such a way that the
body delineation maximizes an object recognition functional $F$. Only
the torso translates over a new search image, while the limbs and head
are carried along during the body search.

Let $\vec{p}$ denote the current search position in image
coordinates. The search for the torso, for example, consists of
projecting the cloud $O_{\lb{Torso}}$ over $I$, by setting the cloud's
current centroid $\vec{c}_{\lb{Torso}}=\vec{p}$, and running a
delineation algorithm $A$ on the set of projected pixels from the
uncertainty region ${\cal U}_{\lb{Torso}}\subseteq I$. Then,
functional $F$ evaluates the set of pixels ${\cal
  M}_{\lb{Torso}}\subseteq I$ labeled by $A$ in $L$ as $\lb{Torso}$
and attributes a score $F_{\lb{Torso}}$ regarding the likelihood of
${\cal M}_{\lb{Torso}}$ actually corresponding to that body
part. However, since we are dealing with 2D projections of a
three-dimensional articulated body in video, changes in pose, zoom,
and rotation require more than simple translation to ensure that each
cloud's uncertainty region ${\cal U}_l$ be properly positioned over
the body part's real boundary in $I$.

We must find the affine transformation $T_l$, for each cloud $O_l\in
{\cal O}$, such that the projection of $O_l$ over $I$ achieves the
best delineation of body part $l$. For such purpose, we first
constrain the search space for $T_l=(\sy_l,\sx_l,\theta_{lk})$ by
defining a set of displacement bounds $\Delta T_l=(\Delta \sy_l,\Delta
\sx_l,\Delta \theta_{lk})$ for the scales $\sy_l,\sx_l$ of cloud/node
$O_l$ and for the relative angle $\theta_{lk}$ of the corresponding
parent joint $e_{lk}\in{\cal E}$. Then, we optimize the affine
transformation parameters for each $T_l$ through Multi-Scale Parameter
Search~\cite{Chiachia11-TR-IC} (MSPS), using the recognition
functional score $F_l$ as the evaluation criterion.

\begin{figure*}[htb]
\begin{center}
\includegraphics[width=\textwidth]{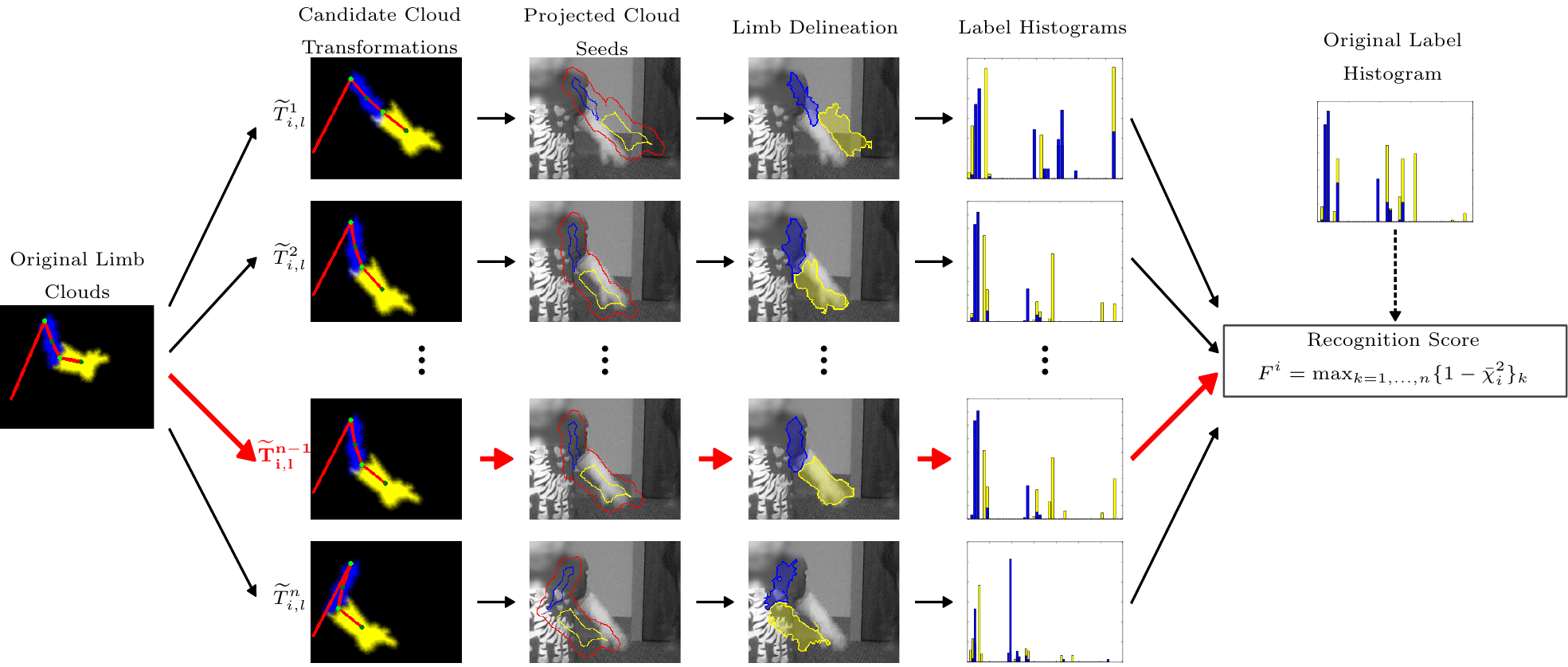} 
\end{center}
\caption{CSM search process of a body limb using Multi-Scale Parameter
  Search (MSPS). Each candidate affine transformation
  $\tcand{T}_{i,l}$ provides a new configuration for the clouds
  $O_{i,l}$ of limb $i$. Then, the seeds of clouds in $i$ are
  projected onto the search frame to delineate all limb parts
  simultaneously. Color histograms are computed for each label $l$ in
  limb $i$, and the mean $\chi^2$ distance to the original histograms
  (from frame $\F{I}{\ti}$) assigns a recognition functional score
  $\bar{F}^i=1-\bar{\chi}^2_i$ to the candidate delineation. MSPS
  maximizes this score to find the projections of clouds $O_{i,l}$
  that best segment limb $i$ (red
  arrows). \label{f.msps-search-diagram}}
\end{figure*}

The MSPS algorithm looks for the optimal parameters of $T_l$ by
searching the solution space $T_l\pm\Delta T_l$ in a gradient descent
fashion, using multiscale steps for each parameter in order to try to
scape from local maxima. For every parameter configuration tested for
$T_l$ during MSPS, cloud $O_l$ is properly transformed according to
the candidate solution $\tcand{T}_l$ and the
projection-delineation-evaluation sequence occurs
(Figure~\ref{f.msps-search-diagram}). The translation of CSM $C$ over
the search image $I$ is easily obtained by adding
$\vec{c}_{\lb{Torso}}$ and $\Delta \vec{c}_{\lb{Torso}}$ to
$T_{\lb{Torso}}$ and $\Delta T_{\lb{Torso}}$, respectively. Since the
search for groups of clouds has shown to be more effective than purely
hierarchical search~\cite{Miranda10-TR-IC}, we conduct the rest of the
body search per branch/limb of $G$, once the optimal parameter
configurations for $T_{\lb{Torso}}$ and $T_{\lb{Head}}$ have been
determined.


The body parts $l$ of limb $i$ are searched simultaneously, by
projecting the clouds $O_{i,l}$ onto $I$ and executing the delineation
algorithm $A$ constrained to the projected pixels ${\cal U}^i\subseteq
I$ (Figure~\ref{f.msps-search-diagram}), where ${\cal U}^i$ is the
combination of the uncertainty regions ${\cal U}_l$ of clouds
$O_{i,l}$ (more details in Section~\ref{ss.csm-delineation}). MSPS
optimizes the affine transformations $T_{i,l}$ for limb $i$,
evaluating the mean object recognition score $\bar{F}^i$ among the
corresponding body parts $l$ of $i$.

The key to positioning the limb clouds $O_{i,l}$ simultaneously is to
allow coordinated changes in their primary axes' scale $\sy_l$ and
parent joint angle $\theta_{lk}$. Hence, the joint displacement vector
$\vec{d}_{lk}$ for joint $e_{lk}\in{\cal E}$ is altered whenever there
is a modification in the scale $\sy_k$ of parent node $k$ or in the
relative angle $\theta_{kP(k)}$ of parent joint $e_{kP(k)}$, thus
moving node/cloud $l$ in the process. Similarly, the centroid
displacement vector $\vec{c}_l$ also changes accompanying the scale
$\sy_l$ of node $l$ and the relative angle $\theta_{lk}$ of edge
$e_{lk}$ (Figure~\ref{f.msps-search-diagram}).

Since MSPS optimizes the parameters of all $T_{i,l}$ for limb $i$, all
scale and angle changes occur simultaneously in order to try a full
range of poses during the search and segmentation of limb
$i$. Notwithstanding, to overcome minor mispositioning of the torso we
allow translation of joints from body parts/clouds directly connected
it (e.g., the neck joint and shoulders). Note that by allowing changes
in the secondary scale $\sx_l$ of all clouds $O_l\in{\cal O}$ we aim
at coping with projective transformations.


The optimal configuration for $G$ of $C$ in image $I$ is simply the
result of hierarchically transforming the clouds of $C$ by $T_l$. We
discuss the selection of the displacement bounds $\Delta T_l$ and
initial search parameters for all clouds in
Section~\ref{s.body-pose-search} (Figure~\ref{f.framework-diagram}).

\subsection{Delineation Algorithm}
\label{ss.csm-delineation}

Our delineation algorithm $A$ works in two steps to achieve
pixel-level delineation of the body in a search image $I$. First, it
outputs a superpixel segmentation mask $R(x)$~\cite{Grundmann10} of
search image $I$ (Figure~\ref{f.framework-diagram}). Then, for every
cloud $O_l\in {\cal O}$ positioned according to the current
configuration of $G$, $A$ simultaneously selects the superpixels of
$R$ completely contained within $O_l$, and partitions the superpixels
that are divided between the cloud's interior, exterior, and
uncertainty regions.

The partitioning of superpixels by algorithm $A$ uses the IFT-SC (IFT
segmentation with Seed Competition), which is based on the Image
Foresting Transform~\cite{Falcao04a} --- a generalization of the
Dijkstra's algorithm that works for multiple sources and \emph{smooth}
path-cost functions. Given the narrow bandwitdth of the uncertainty
regions, any delineation algorithm would provide similar results to
IFT-SC (e.g., graph cuts~\cite{Boykov06}, fuzzy
connectedness~\cite{Udupa02}, random walks~\cite{Grady06}, and power
watershed~\cite{Couprie11}). Nevertheless, IFT-SC has proven to
provide equivalent solutions to graph cuts~\cite{Miranda09c} and fuzzy
connectedness~\cite{Ciesielski12} under certain conditions, while
handling multiple objects simultaneously in linear time over the
number of pixels of the uncertainty regions~\cite{Falcao04a}. For a
comparison between IFT-SC and other algorithms,
see~\cite{Ciesielski12}.

IFT-SC considers the image graph $\{I, {\cal A}_8\}$ with all the
pixels $x\in I$ being the nodes, and an adjacency relation ${\cal
  A}_8$ connecting every $8$-neighbor pixel in $I$. A \textit{path}
$\pi=\langle z_1,z_2,\ldots,z_n \rangle$ is a sequence of adjacent
nodes in the image graph. A \emph{connectivity function} $f$ assigns a
path-cost value $f(\pi)$ to any path $\pi$ in $I$. We consider the
following connectivity function
\begin{eqnarray}
f_\eta(\pi) & = & \left\{\begin{array}{ll}
\sum^{n-1}_{j=1}[w(z_j,z_{j+1})]^\eta & \mbox{if } z_1\in {\cal S},
\\ +\infty & \mbox{otherwise,}
\end{array}\right.
\label{e.path-cost}
\end{eqnarray}
where $w(z_j,z_{j+1})$ is a weight for arc $(z_j,z_{j+1})\in {\cal
  A}_8$, and ${\cal S}$ is a set of specially selected pixels denoted
as \emph{seeds} (Figure~\ref{f.msps-search-diagram}). The superpixels
from $R$ usually follow the image edges properly
(Figure~\ref{f.framework-diagram}), but some superpixels contain
pixels from both the foreground and background regions, which must be
separated. Hence, we define the arc weight
$w(z_j,z_{j+1})=\frac{|\nabla I(z_j)|+|\nabla I(z_{j+1})|}{2}$
considering the mean magnitude of the image gradient $|\nabla I(z)|$
(computed from \emph{Lab} color differences) of pixels $z_j$ and
$z_{j+1}$.\footnote{A gradient of the cloud image $O_l$ may also be
  combined with the arc weights to fill missing gaps of $\nabla
  I$~\cite{Miranda10-TR-IC}. In this case, it is interesting to
  previously narrow the uncertainty region ${\cal U}_l$ by adjusting
  the parameters of Eq.~\ref{e.sigmoid}.}

Being $r$ a superpixel of $R$ completely contained inside the interior
region of $O_l$ projected over $I$, we can straightforwardly assign
label $L(x)=l, \forall R(x)=r$. If we have instead $0\leq O_l(z)<1$
for some pixels such that $R(z)=r$, each pixel $z$ must be labeled
according to how strongly connected it is to either $O_l$ or the
background. Let ${\cal S}^f_l$ and ${\cal S}^b_l$ denote the sets of
seed pixels from the interior (foreground) and exterior (background)
of cloud $O_l$, respectively, being on the boundary of the uncertainty
region of $O_l$ with at least one $8$-neighbor pixel in ${\cal
  U}_l$. Seed sets ${\cal S}^f_l$ and ${\cal S}^b_l$ compete for the
pixels of the uncertainty region ${\cal U}_l\subseteq I$ projected
onto $I$ by defining ${\cal S}={\cal S}^f_l\cup{\cal S}^b_l$ in
Eq.~\ref{e.path-cost}, such that $z$ receives label $L(z)=0$ if the
minimum-cost path comes from a seed in ${\cal S}^b_l$ and $L(z)=l$,
otherwise. We constrain the competition according to the superpixels
of $R$, by allowing paths in the graph to exist only between
neighboring pixels $(z_j,z_{j+1})\in {\cal A}_8$ where
$R(z_j)=R(z_{j+1})$. The delineation of body part $l$ is then defined
as the union between the interior of the cloud $O_l$ and the pixels
with labels $L(x) = l$ in ${\cal U}_l$. Note that, for each limb $i$
the seed set ${\cal S}$ in Eq.~\ref{e.path-cost} includes the seeds of
all clouds $O_{i,l}$, which compete simultaneously for the union of
the projected uncertainty regions ${\cal U}^i\subseteq I$
(Figure~\ref{f.msps-search-diagram}).\footnote{We prevent
  superimposition of clouds by eliminating seeds $x\in{\cal S}^b_l$ if
  $O_{i,h}(x)=1$ for any cloud from $i$ such that $h\neq l$.}

The IFT-SC solves the above minimization problem by computing an
\emph{optimum-path forest} --- a function $P$ that contains no cycles
and assigns to each node $z\in I$ either its predecessor node $P(z)\in
I$ in the optimum path with terminus $z$ or a distinctive marker $P(z)
= nil\notin I$, when $\langle z \rangle$ is optimum (i.e., $z$ is said
to be a \emph{root} of the forest). The cost function $f_\eta$ in
Eq.~\ref{e.path-cost} forces the roots to be in ${\cal S}$. By using
the parameter $\eta=1.5$ in $f_\eta$, we obtain more regularization on
the object's boundary~\cite{Miranda10-TR-IC}, as opposed to using the
commonly adopted function for IFT-SC that considers the maximum arc
weight along the path. The IFT-SC delineation is very efficient since
it can be implemented to run in linear time with respect to the size
of the uncertainty region(s) of the cloud(s)~\cite{Falcao04a}, which
in turn is much smaller than $|I|$.

\section{Human Body Search in Video Using the CSM}
\label{s.body-pose-search}

After computing the Cloud System Model $C$ in frame $\F{I}{\ti}$, $C$
is used to search for the toddler in frame $\F{I}{\tc}$ using MSPS,
with $\tc>\ti$. The previous configuration $G^{\tp}$ of $C$ would then
be the starting point for finding the optimal configuration $G^{\tc}$
in the next frame. Since video data is available, temporal information
allows us to look instead for an initial guess that is closer to
$G^{\tc}$ than $G^{\tp}$ (i.e., we ``warp'' the CSM to $\F{I}{\tc}$,
Figure~\ref{f.framework-diagram}). This is done by estimating the set
of parameters for the affine transformations $T^*_l$ (and
corresponding $\Delta T^*_l$) as an initial guess for $T^{\tc}_l$,
from the motion of non-background pixels $x\in\F{L}{\tp}$ to frame
$\F{I}{\tc}$.

\subsection{Initial Search Parameter Estimation}

Let $\F{L}{\tc}^*$ be the propagated label image $\F{L}{\tp}$ to
$\F{L}{\tc}$ using dense optical flow~\cite{Tepper12}
(Figure~\ref{f.framework-diagram}), after applying a median filter to
cope with noise. For every node $l\in G^{\tp}$, estimating changes in
scale of the axes of the cloud $O^{\tp}_l$ in frame $\F{I}{\tc}$
involves first determining the global orientation of $O^{\tp}_l$ in
image coordinates. Again, we assume that the cloud's orientation is
the same of the propagated body part $l\in\F{L}{\tc}^*$ and compute
it using PCA from the labeled pixel coordinates. The initial scales
for the primary and secondary axes of body part $l$ are proportional
to the change in variance of the labeled pixel coordinates, projected
onto the corresponding axes of $l$, between $\F{L}{\tp}$ and
$\F{L}{t+1}^*$.
The estimated relative angle $\theta^*_{lk}$ derives directly from the
global orientations of clouds $O_l$ and $O_k$, for every joint
$e_{lk}\in {\cal E}$. Lastly, the estimated joint displacement vector
$\vec{d}^*_{lk}$ is simply obtained by adding $\vec{d}^{\tp}_{lk}$ to
the median propagation displacements of all pixel coordinates $x\in
\F{L}{\tp}$, such that $\F{L}{\tp}(x)=l$.

Since we already consider the motion propagation to estimate $T^*_l$,
we define the displacement bounds $\Delta T^*_l$ according to our
prior knowledge of the human body's movements. For the limb joints'
relative angles we allow them to move $\Delta \theta_{lk} =
30^o$. Similarly, we constrain the neck joint angle to move
$5^o$. Changes in scale can be at most $2\%$, while we set $\Delta
\vec{d}^*_{lk}=\beta \cdot |\vec{d}^{\tp}_{lk}-\vec{d}^*_{lk}|$ to
allow the joints for body parts linked to the $k=\lb{Torso}$ to move
proportionally to the part's estimated motion ($\beta=1.5$). The same
parameters also apply to preventing sudden limb motions, which
characterize erroneous motion estimation. These impositions can be
further improved if we exploit physics-based kinematic models of the
human muscle structure~\cite{Sherman11}.

\subsection{Object Recognition Functional}
\label{ss.csm-functional}

The last part of our method that needs to be defined for finding the
toddler's body in frame $\F{I}{\tc}$ using MSPS is the recognition
functional $F$ of $C$. $F$ takes into account the comparison of color
histograms across frames to ouput a score for the delineation result
during the body search using MSPS. More precisely, color histograms
are computed for the pixels ${\cal M}^{\ti}_l\subseteq \F{L}{\ti}$ of
every body part in frame $\F{I}{\ti}$, considering the quantized RGB
colorspace ($16$ bins per channel). These histograms are redefined
after each search delineation in frame $\F{I}{\tc}$ using the object
labeled pixels by the IFT-SC. Then, the recognition functional score
for the current search position is $F_l=1-\chi^2_l$, the complement of
the $\chi^2$ distance between the histograms of frames $\F{I}{\ti}$
and $\F{I}{\tc}$, for each body part $l$
(Figure~\ref{f.msps-search-diagram}) --- we evaluate the mean
recognition score $\bar{F}^i$ among the parts of limb $i$ when
searching for it.

After the toddler's body is properly found and segmented in frame
$\F{I}{\tc}$, the resulting segmentation label $\F{L}{\tc}$ and pose
configuration given by $G^{\tc}$ are used to reestimate the search
parameter for frame $\F{I}{\tn}$
(Figure~\ref{f.framework-diagram}). We keep the histograms from the
first frame $\F{I}{\ti}$ for comparison in frame $\F{I}{\tc}$, where
$\tc>\ti$, for greater stability~\cite{Bai09b}.

\subsection{Body Pose From the Relational Model}

The toddler's body pose in $\F{I}{\tc}$ can be straightforwardly
obtained from the joint configuration of $G^{\tc}$ in image
coordinates. The only care that must be taken is when the hands (or
feet) are not part of the cloud system. In such situations, instead of
connecting the elbow to the wrist to define the forearm segment, we
compute the skeleton by connecting the elbow to the forearm cloud's
center (Figure~\ref{f.pose-estimation}). Afterwards, we use the
skeleton to determine arm symmetry at time $\tc$
(Section~\ref{ss.asymmetry-measurement}).

\section{Aiding Autism Assessment}
\label{s.autism-assessment}

Motor development disorders are considered some of the first signs
that could preclude social or linguistic abnormalities~\cite[and
  references therein]{Esposito11}. Detecting and measuring these
atypical motor patterns as early as in the first year of life can lead
to early diagnosis, allowing intensive intervention that improves
child outcomes~\cite{Dawson08}. Despite this evidence, the average age
of ASD diagnosis in the U.S. is 5 years~\cite{Shattuck09}, since most
families lack easy access to specialists in ASD. There is a need for
automatic and quantitative analysis tools that can be used by general
practitioners in child development, and in general environments, to
identify children at-risk for ASD and other developmental
disorders. This work is inserted in a long-term multidisciplinary
project~\cite{Hashemi12a,Hashemi12b,Fasching12} with the goal of
providing non-intrusive computer vision tools, that do not induce
behaviors and/or require any body-worn sensors (as opposed
to~\cite{Goodwin11,Nazneen10}), to aid in this early detection
task.\footnote{Behavioral Analysis of At-Risk Children,
  website:\emph{http://baarc.cs.umn.edu/}}

\begin{figure}
\begin{center}
\begin{tabular}{cc}
\includegraphics[width=.25\textwidth]{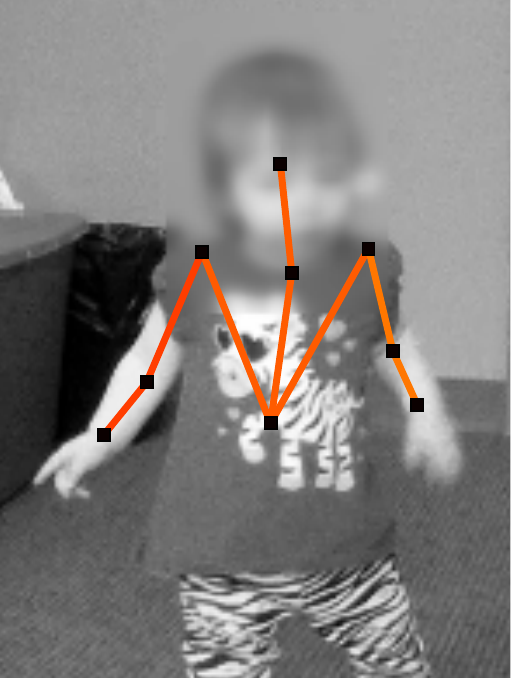} &
\includegraphics[width=.25\textwidth]{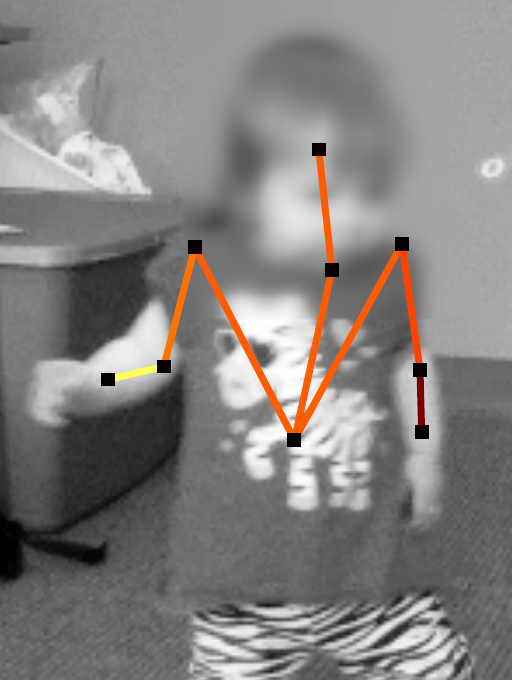}
\end{tabular} 
\caption{Example of symmetric and asymmetric arms. The sticks
  (skeleton) are automatically positioned with our technique.
\label{f.asymmetric-arm}}
\end{center}
\end{figure}

Children diagnosed with autism may present arm-and-hand flapping, toe
walking, asymmetric gait patterns when walking unsupportedly, among
other stereotypical motor behaviors. In particular, Esposito et
al.~\cite{Esposito11} have found that diagnosed toddlers often
presented asymmetric arm positions (Figure~\ref{f.asymmetric-arm}),
according to the Eshkol-Wachman Movement Notation
(EWMN)~\cite{Teitelbaum04}, in home videos filmed during the
children's early life period. EWMN is essentially a 2D stickman that
is manually adjusted to the child's body on each video frame and then
analyzed. Symmetry is violated, for example, when the toddler walks
with one arm fully extended downwards alongside his/her body, while
holding the other one horizontally, pointing forward
(Figure~\ref{f.asymmetric-arm}). Performing such analysis is a
burdensome task that requires intensive training by experienced
raters, being impractical for clinical settings. We aim at
semi-automating this task by estimating the 2D body pose of the
toddlers using the CSM in video segments in which they are walking
naturally.

As an initial step towards our long-term goal, we present results from
actual clinical recordings, in which the at-risk infant/toddler is
tested by an experienced clinician using a standard battery of
developmental and ASD assessment measures. The following subsection
describes how we compute arm asymmetry from the by-product skeleton of
the CSM segmentation. Then, we present results obtained from our
clinical recordings that can aid the clinician in his/her assessment.

\subsection{Arm Asymmetry Measurement}
\label{ss.asymmetry-measurement}

\newcommand{\ASa}[0]{\ensuremath{AS^*}}
\newcommand{\ASu}[0]{\ensuremath{AS_u}}
\newcommand{\ASf}[0]{\ensuremath{AS_f}}
\newcommand{\ADf}[0]{\ensuremath{AD_f}}

Following~\cite{Esposito11}, a symmetrical position of the arms is a
pose where similarity in relative position of corresponding limbs (an
arm and the other arm) is shown with an accuracy of $45^o$. This
happens because EWMN defines a 3D coordinate system for each body
joint that discretizes possible 2D skeleton poses by dividing the 3D
space centered at the joints into $45^o$ intervals.

From our dataset, we have seen that using simple measures obtained
directly from the 2D skeleton is often insightful enough to detect
most cases of arm asymmetry, thus avoiding the manual annotation
required by EWMN according to the aforementioned coordinate
system. For such asymmetry detection task, we define the following
normalized asymmetry score for each arm segment:
\begin{eqnarray}
AS & = & \frac{2.0}{1.0+\exp{(-\frac{\alpha - \tau}{\sigma_{\tau}}})}
\label{e.seg-asymmetry-score},
\end{eqnarray}
where $\alpha$ is the absolute difference between either global or
relative 2D angles obtained from corresponding left/right arm
segments, $\tau$ is a given asymmetry threshold, and $\sigma_\tau$ is
a parameter set to control acceptable asymmetry values. Considering
EWMN's accuracy, we set the asymmetry threshold $\tau=45^o$. We have
empirically observed that $\sigma_\tau = \frac{\tau}{3}$ helps coping
with near asymmetrical poses when outputing the asymmetry score.

\begin{figure}[htb]
\begin{center}
\def\svgwidth{.15\textwidth}
\includegraphics[width=.35\textwidth]{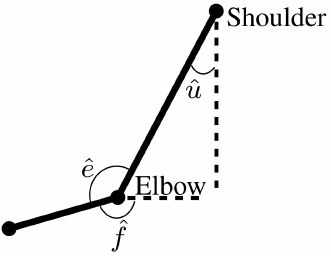} 
\end{center}
\caption{Angles used to compute the arm asymmetry scores.
  \label{f.as-angles}}
\end{figure}

For the upper arm asymmetry score $\ASu$, $\alpha$ in
Eq.~\ref{e.seg-asymmetry-score} is the absolute difference
$\alpha=|\hat{u}_l-\hat{u}_r|$ between the global angles $\hat{u}_l$
and $\hat{u}_r$ formed between the left and right upper arms and the
vertical axis, respectively (Figure~\ref{f.as-angles}). The forearm
asymmetry score $\ASf$ is similarly defined by setting
$\alpha=|\hat{e}_l - \hat{e}_r|$, where $\hat{e}$ is the relative
forearm angle with respect to the upper arm formed by the elbow
(Figure~\ref{f.as-angles}). The asymmetry score for the entire arm is
finally defined as $\ASa = \max{\{\ASu,\ASf\}}$.

The rationale behind $\ASa$ is that if the toddler's upper arms are
pointing to different (mirrored) directions, then the arms are
probably asymmetric and $\ASu$ should be high (i.e.,
$\ASa\geq1.0$). Otherwise, if $\ASf$ is great then one arm is probably
stretched while the other one is not, thus suggesting arm
asymmetry. Regardless, we may also show where the forearms are
pointing to as another asymmetry measure, by analysing their global
angles $\hat{f}_l$ and $\hat{f}_r$ w.r.t. the horizontal axis
(Figure~\ref{f.as-angles}). If the absolute difference
$\ADf=|\hat{f}_l-\hat{f}_r|$ between those global angles is greater
than $45^o$, for example, then the arm poses are probably
asymmetric~\cite{Hashemi12a}. Both $\ASa$ and $\ADf$ have different
advantages and shortcomings that will be discussed in the results
Section~\ref{s.results-skeleton}.

Since we are interested in providing measurements for the clinician,
we output temporal graphs for each video segment with the
aforementioned single-frame asymmetry measures.
From these measurements, different data can be extracted and
interpreted by the specialists. Esposito et al.\cite{Esposito11}, for
instance, look at two different types of symmetry in video sequences:
Static Symmetry (SS) and Dynamic Symmetry (DS). The former assesses
each frame individually, while the latter evaluates groups of frames
in a half-second window. If at least one frame is asymmetric in a
window, then the entire half-second is considered asymmetric for
DS. SS and DS scores are then the percentage of asymmetric frames and
windows in a video sequence, respectively (the higher the number, the
more asymmetrical the walking pattern). Although we do not aim at
fully reproducing the work of~\cite{Esposito11}, we attempt to
quantify asymmetry for each of our video sequences by computing SS and
DS.

\newcommand{\pnineteen}{\ensuremath{\#1}}
\newcommand{\pfive}{\ensuremath{\#2}}
\newcommand{\ptwo}[0]{\ensuremath{\#3}}
\newcommand{\pthree}[1]{%
  \ifthenelse{\isempty{#1}}%
    {\ensuremath{\#4}}
    {\ensuremath{#1 4}}
}
\newcommand{\pone}[0]{\ensuremath{\#5}}
\newcommand{\pseven}[0]{\ensuremath{\#6}}

\newcommand{\graphfigwidth}{.75\textwidth}
\newcommand{\graphwidth}{.32\textwidth}

\newcommand{\figresultwidth}{.12\textwidth}
\newcommand{\figresultheight}{.12\textheight}
\newcommand{\vspfnum}{-6pt}
\newcommand{\vspafterfnum}{0pt}

\newcommand{\fnum}[1]{{\scriptsize #1}}

\section{Experimental Validation}
\label{s.results-skeleton}

\begin{figure*}
\begin{center}
\begin{tabular}{SCCCCCC}
  \pnineteen &
  \includegraphics[width=\figresultwidth]{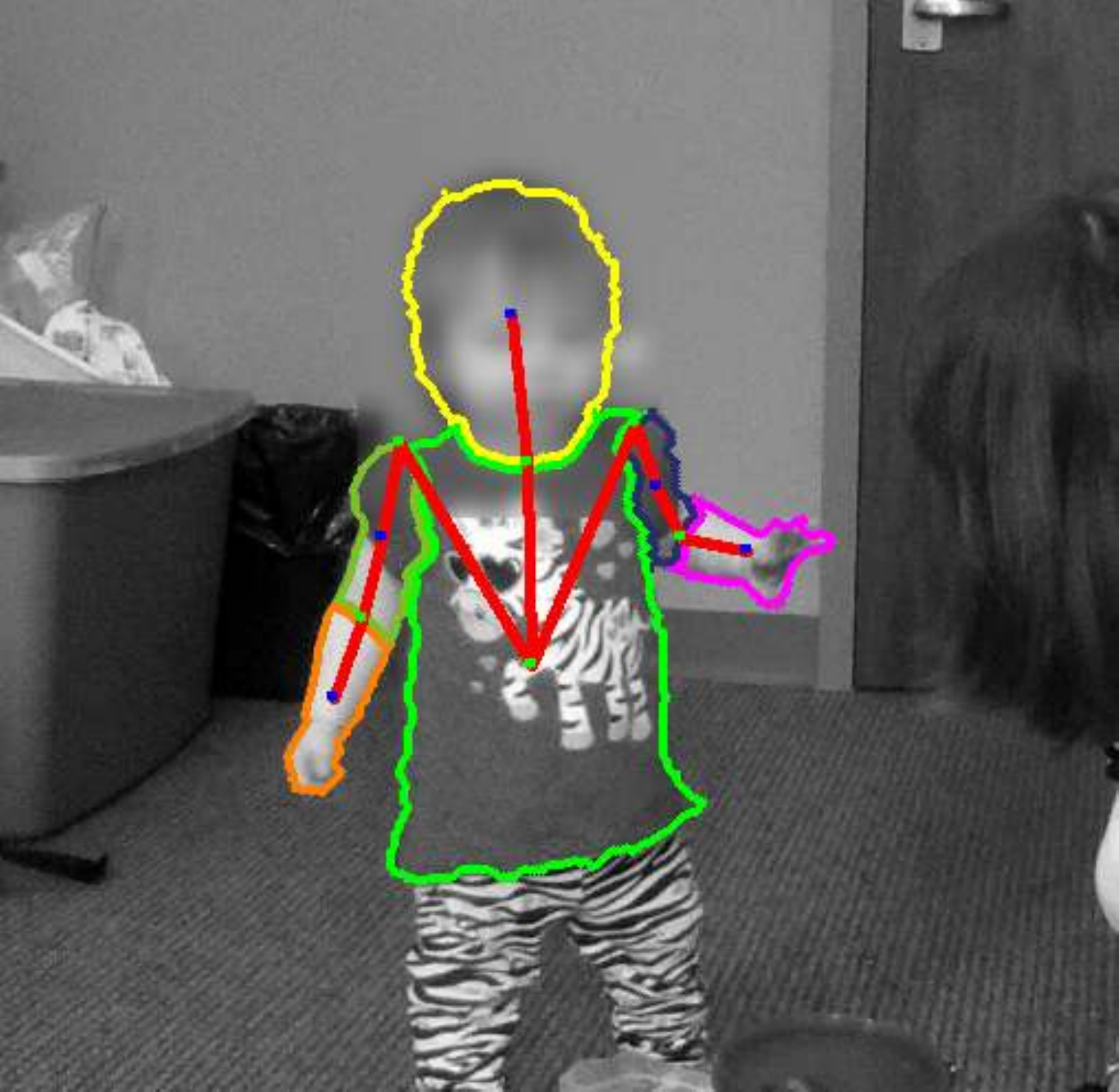} &
  \includegraphics[width=\figresultwidth]{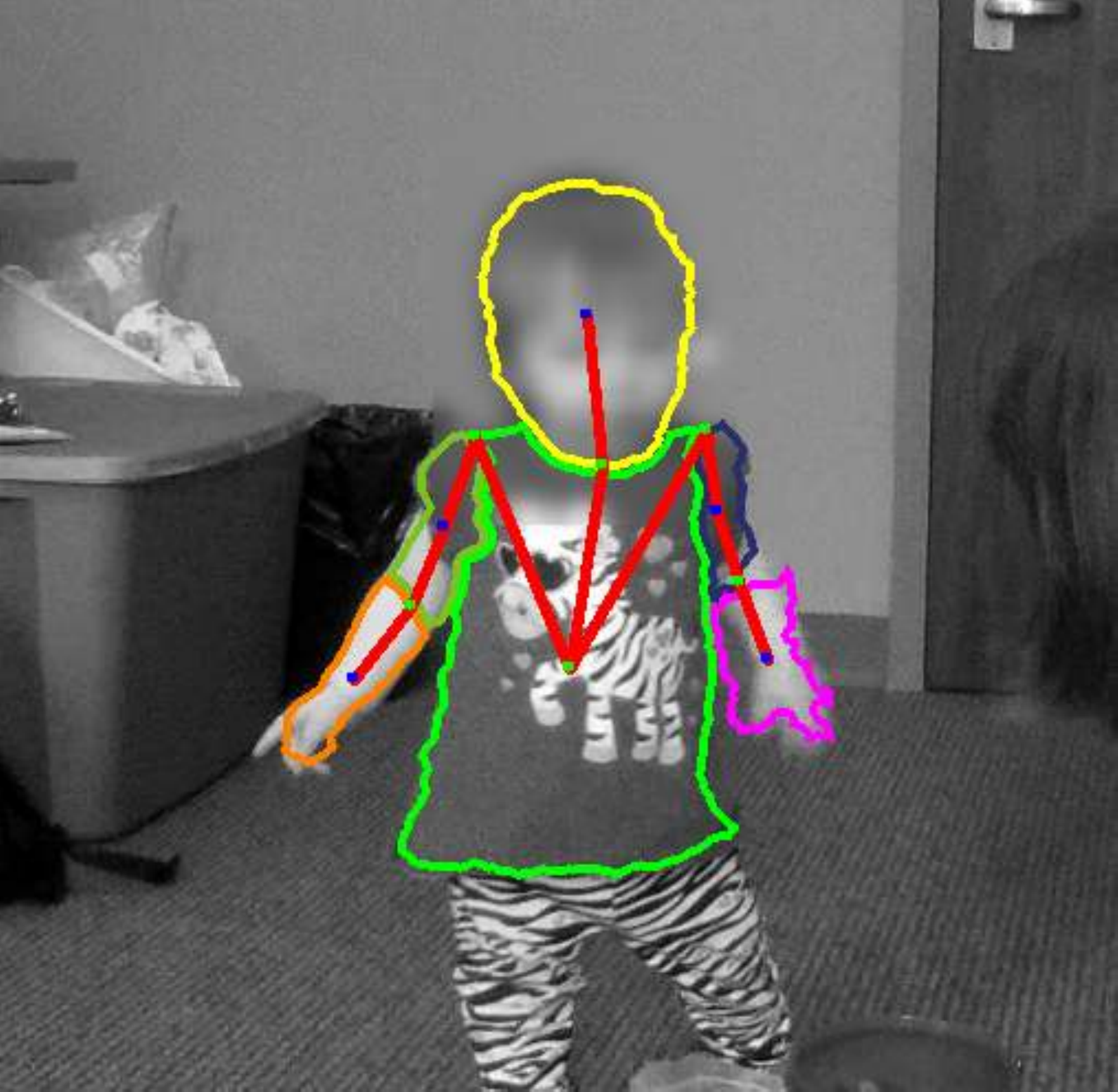} &
  \includegraphics[width=\figresultwidth]{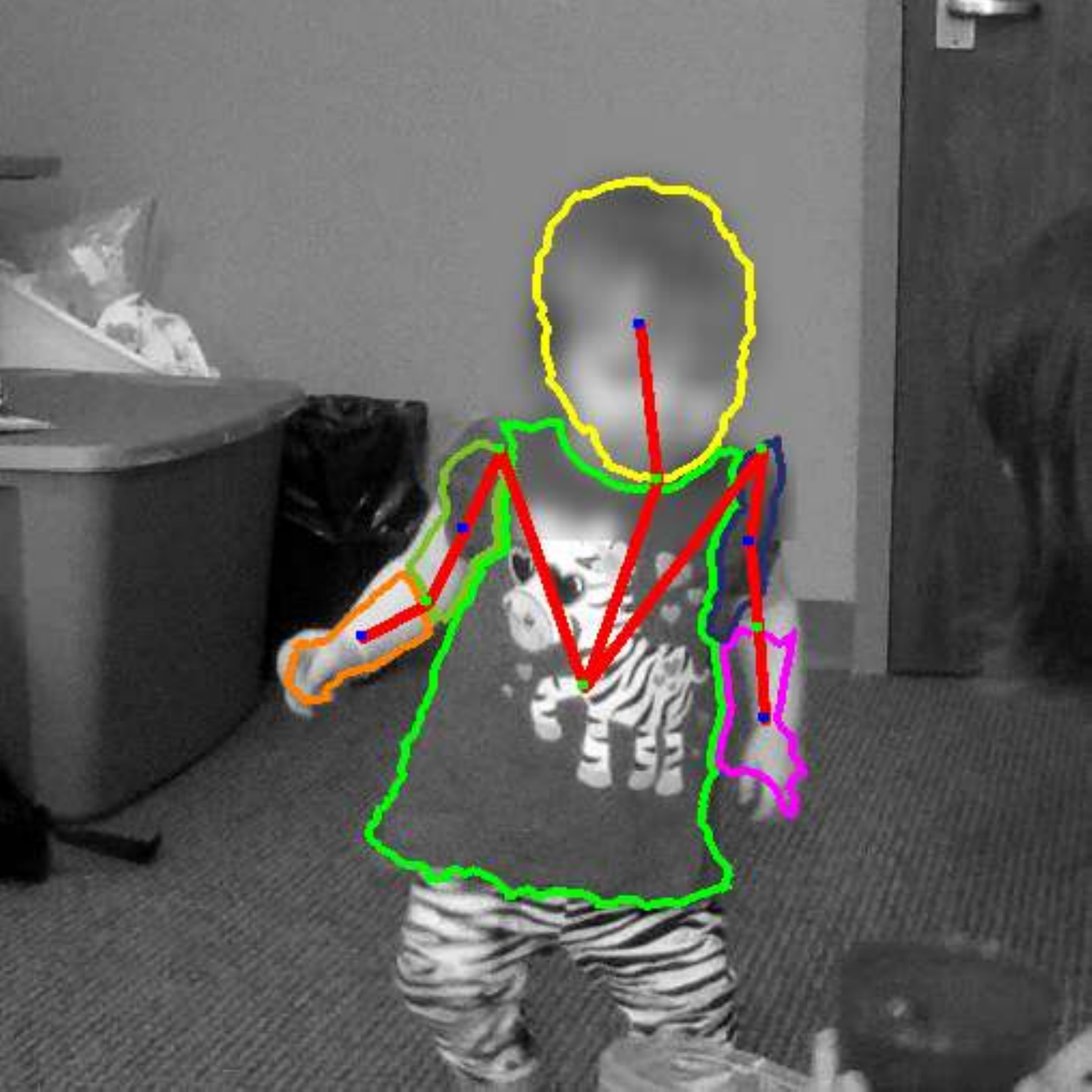} &
  \includegraphics[width=\figresultwidth]{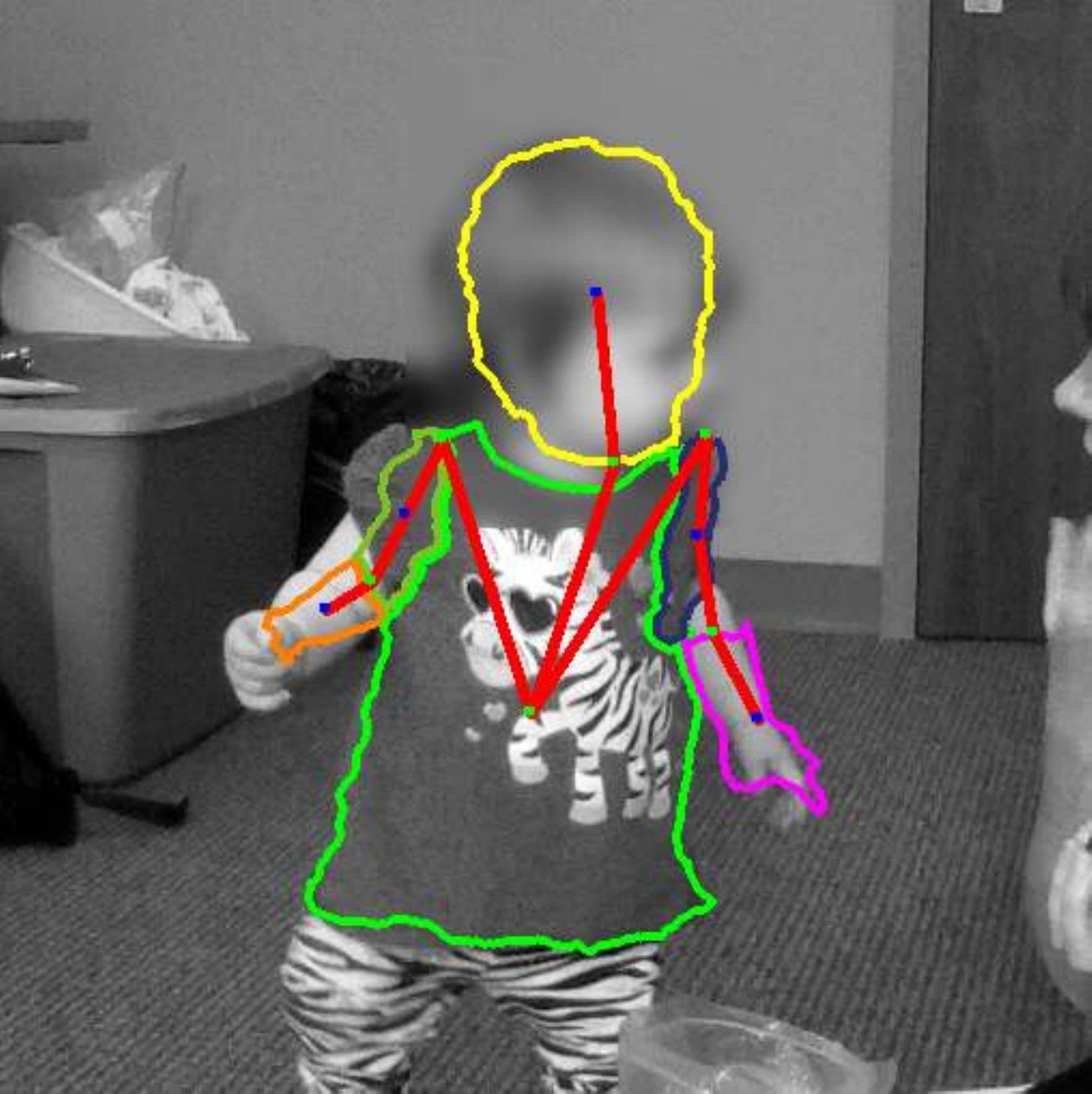} &
  \includegraphics[width=\figresultwidth]{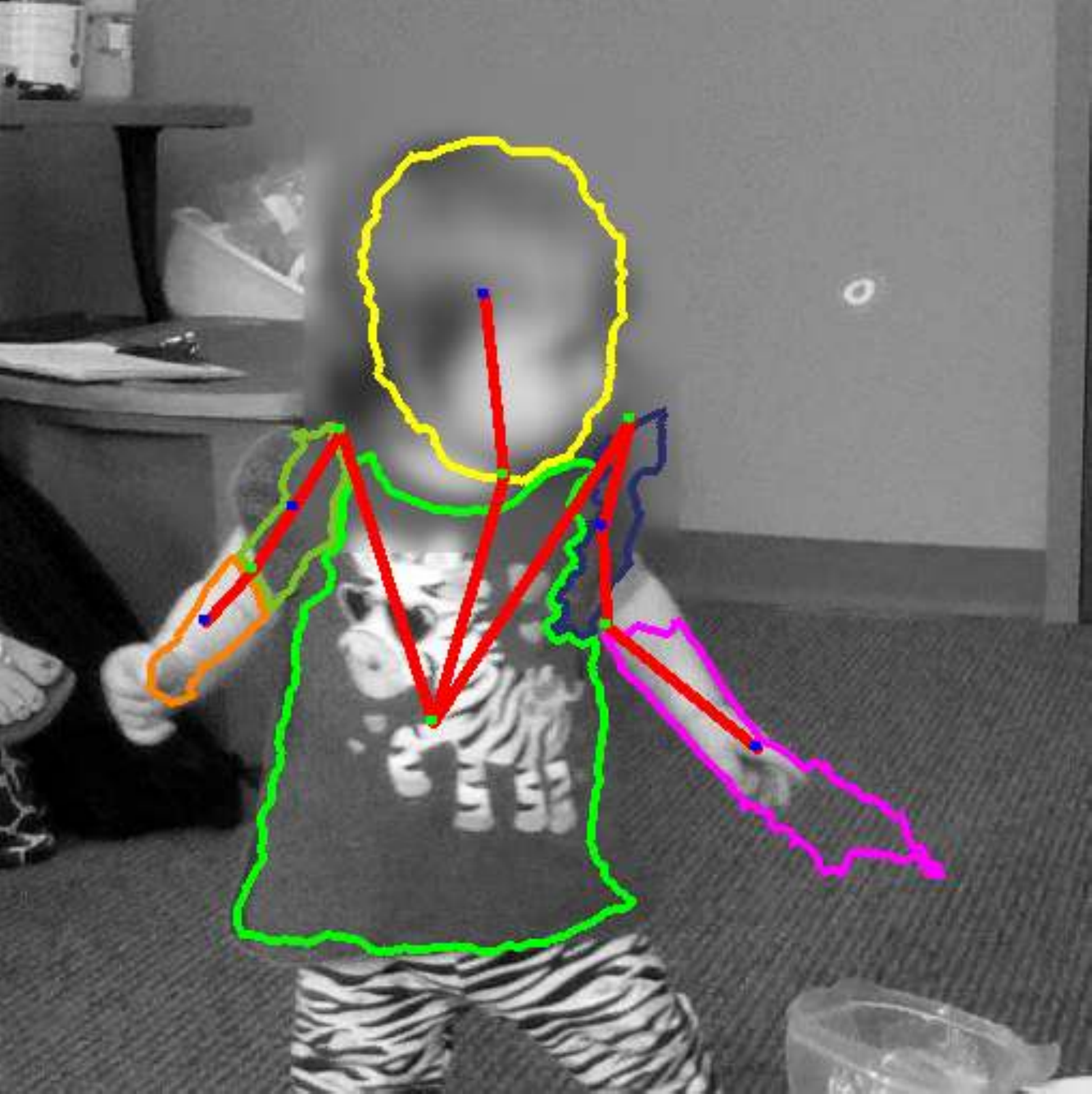} &
  \includegraphics[width=\figresultwidth]{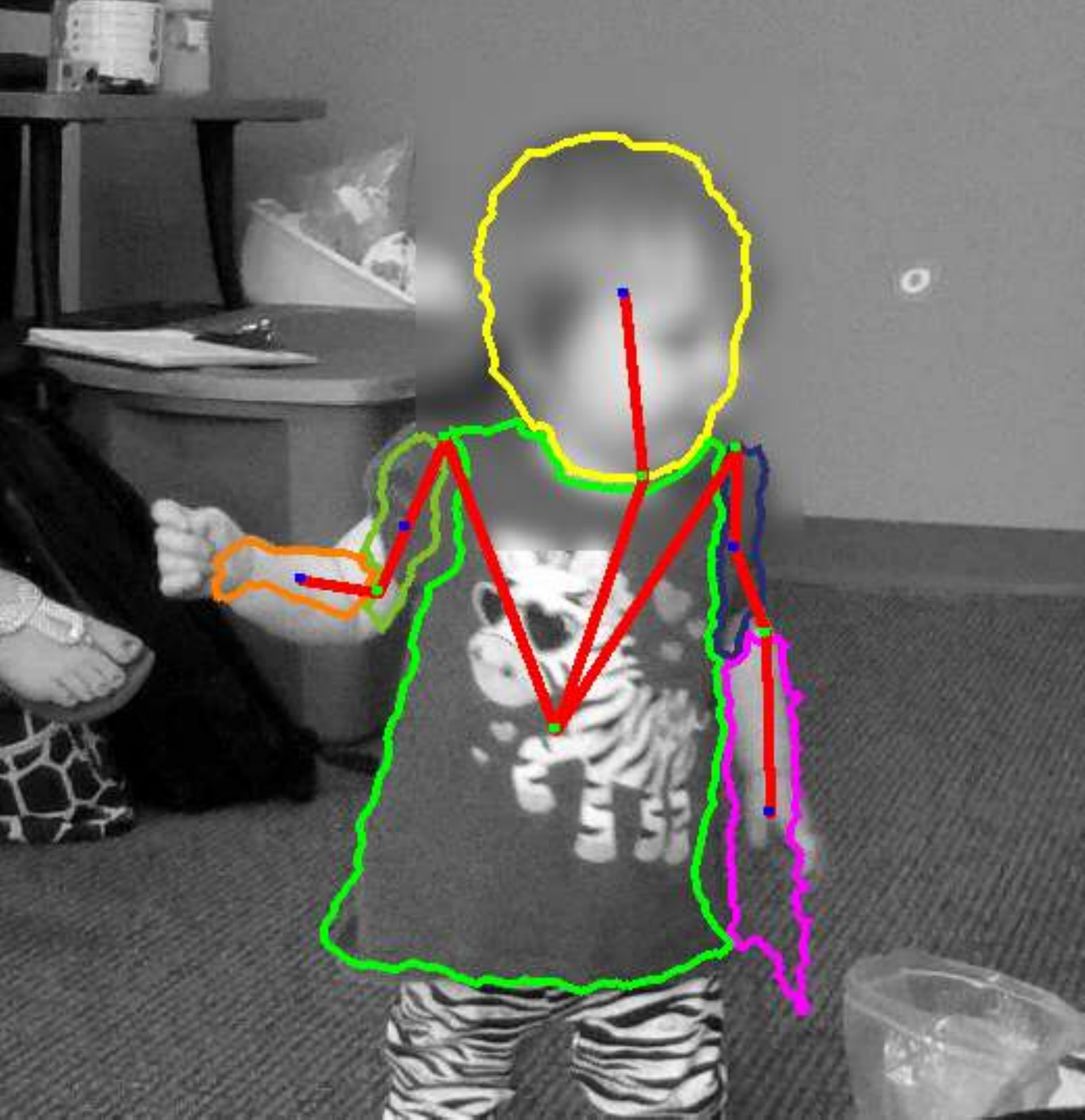}\\ 
  & \fnum{\ti} & \fnum{22} & \fnum{44} & \fnum{88} & \fnum{110} & \fnum{150}\\
  \pfive &
  \includegraphics[width=\figresultwidth,trim=0 0 15px 0,clip=true]{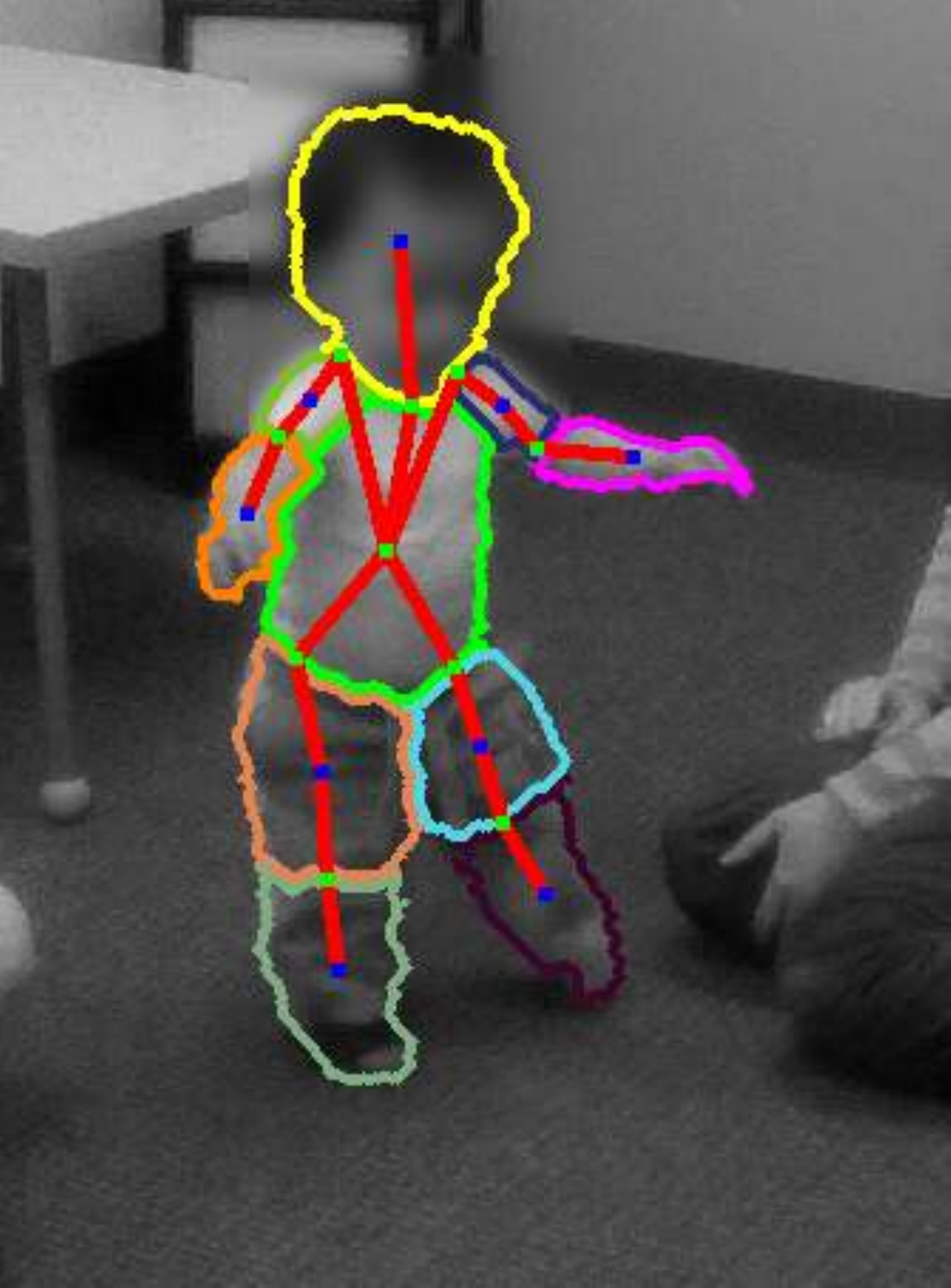} &
  \includegraphics[width=\figresultwidth,trim=0 0 10px 0,clip=true]{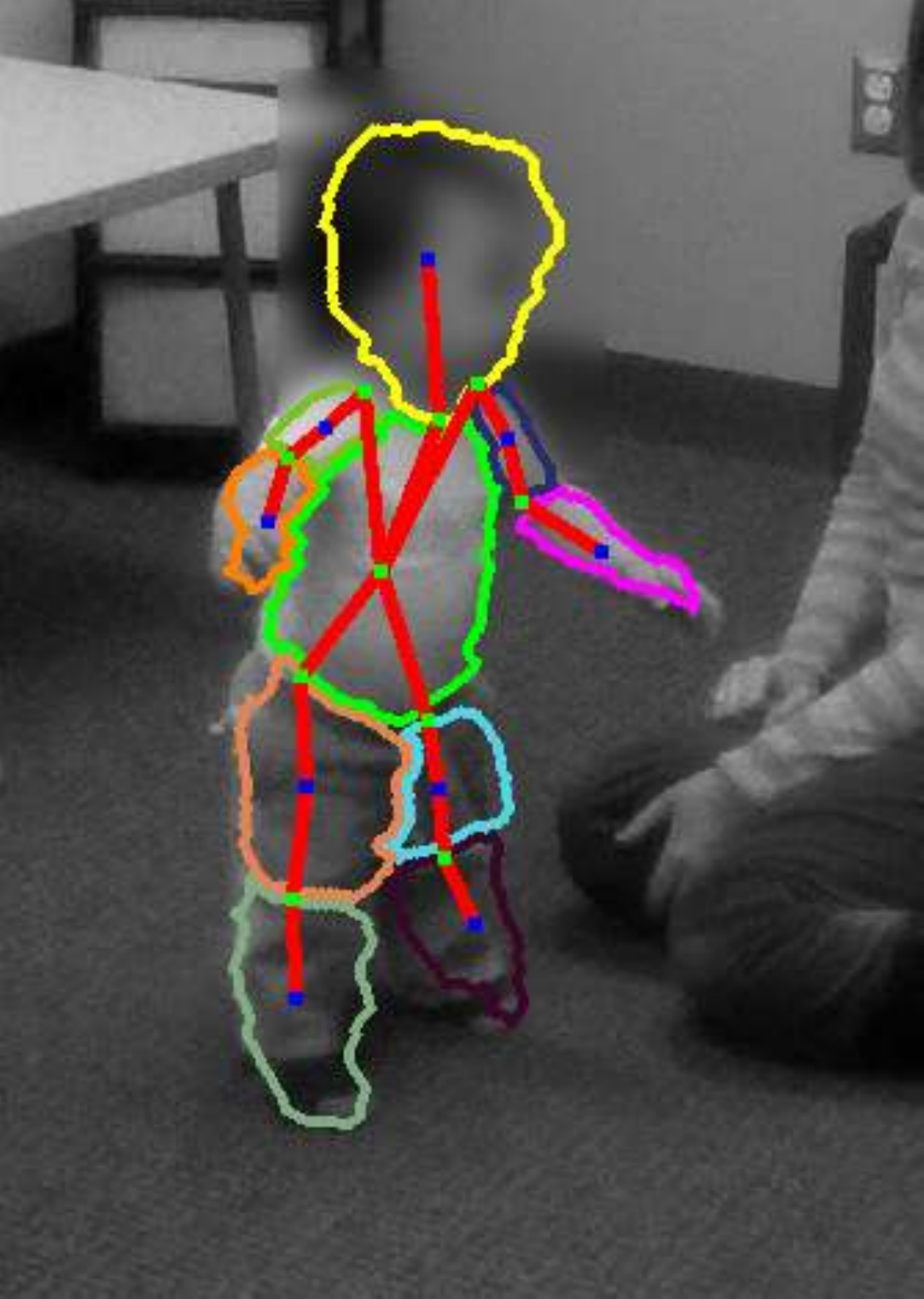} &
  \includegraphics[width=\figresultwidth,trim=0 0 10px 0,clip=true]{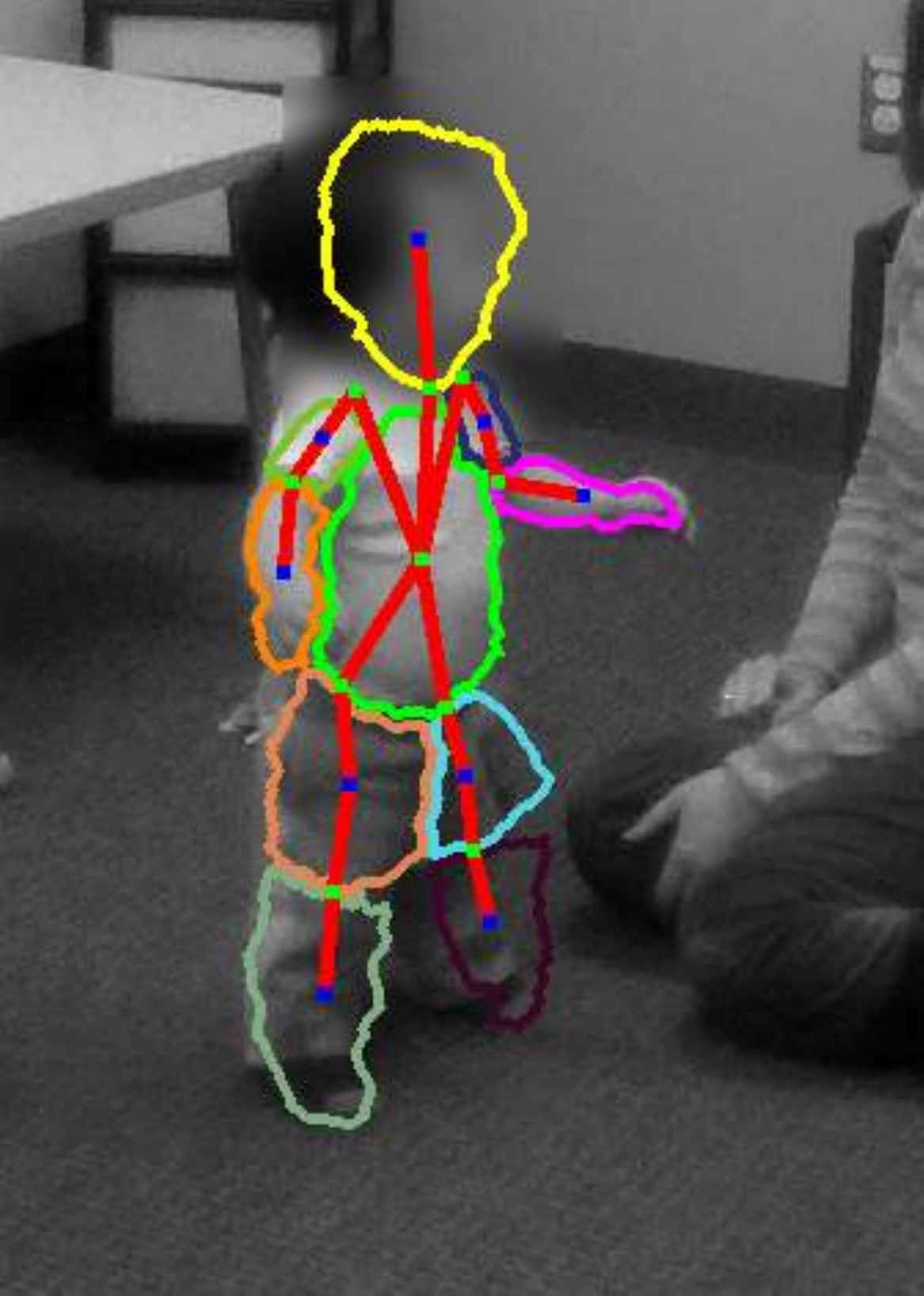} &
  \includegraphics[width=\figresultwidth]{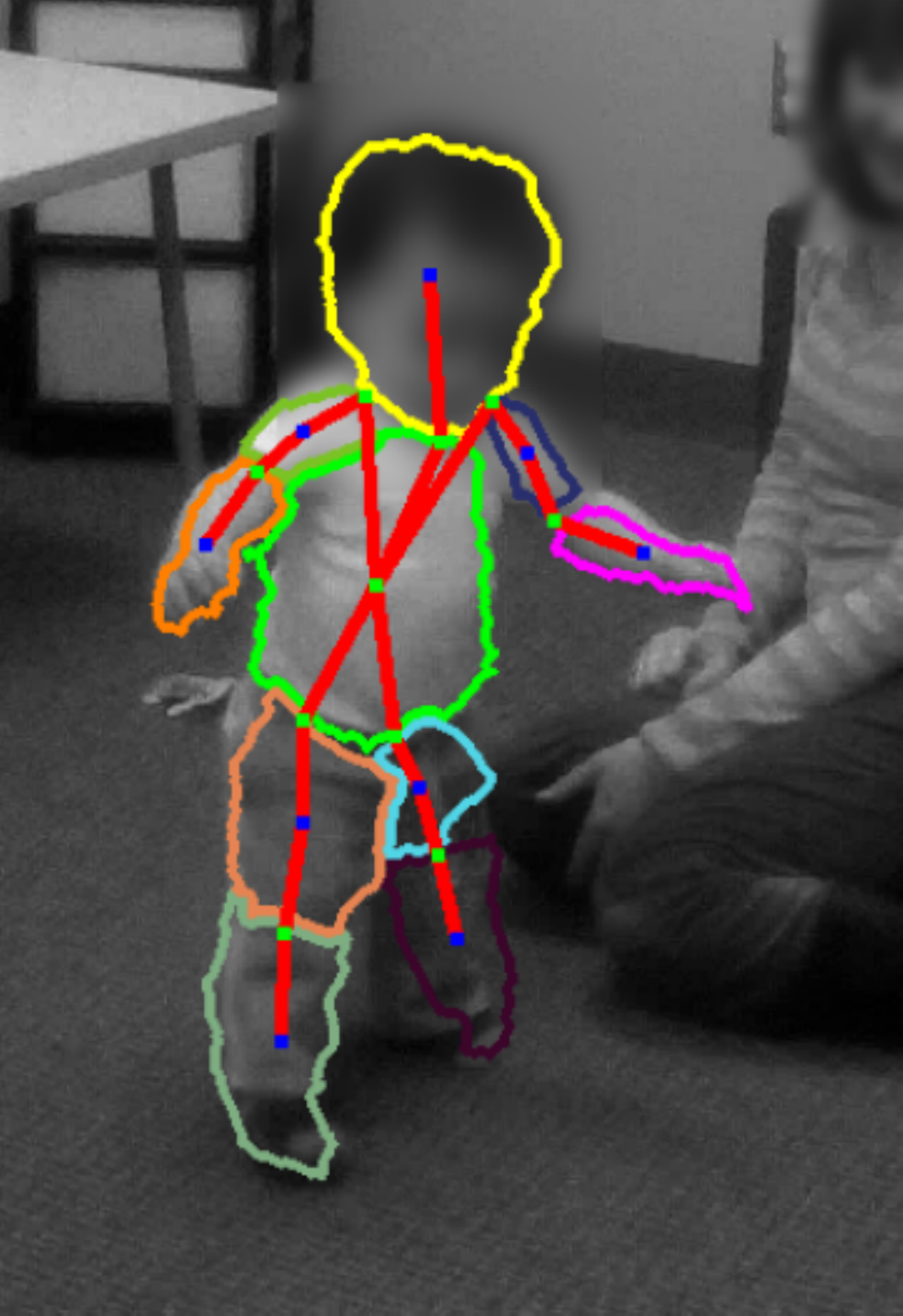} &
  \includegraphics[width=\figresultwidth]{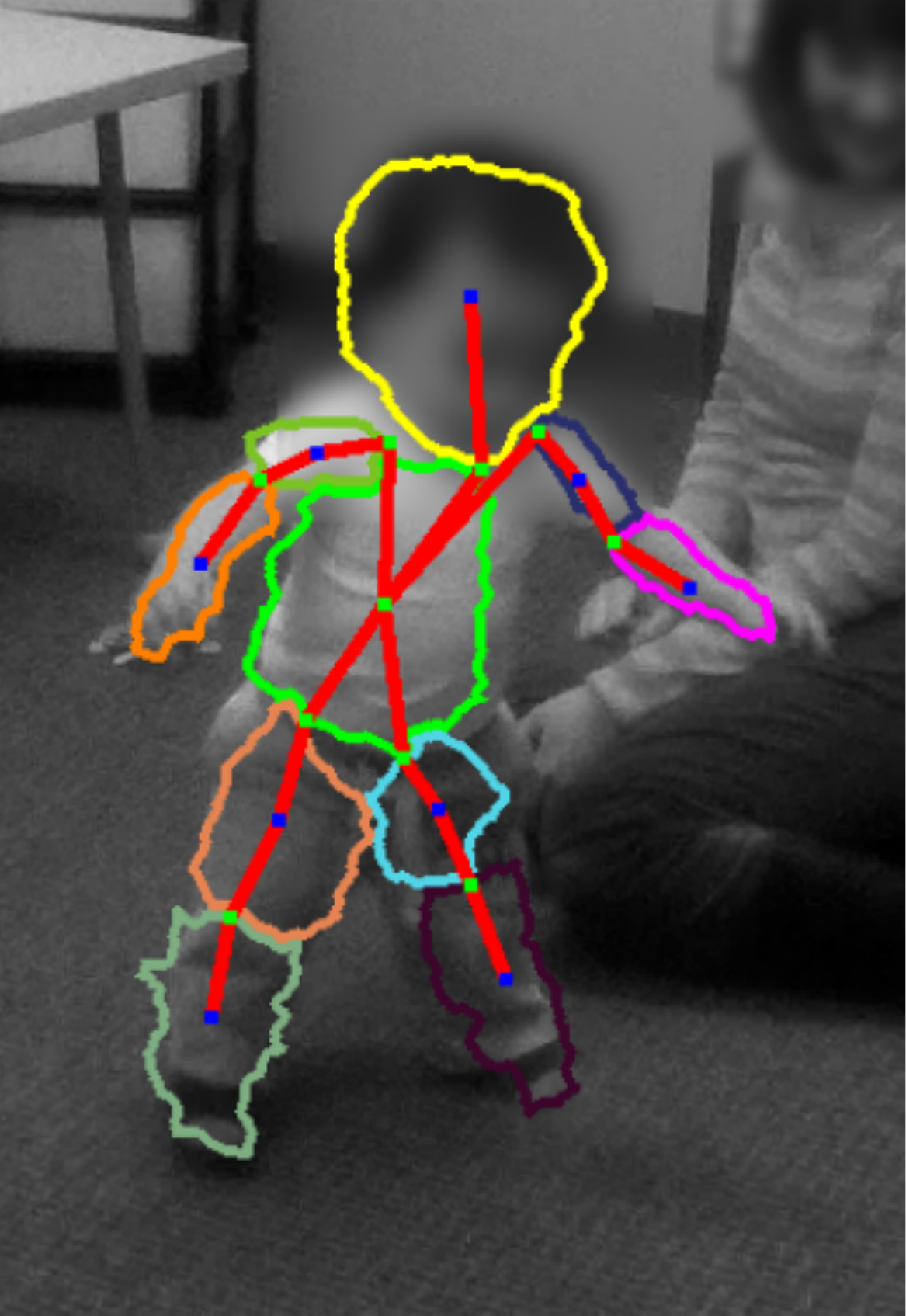} &
  \includegraphics[width=\figresultwidth]{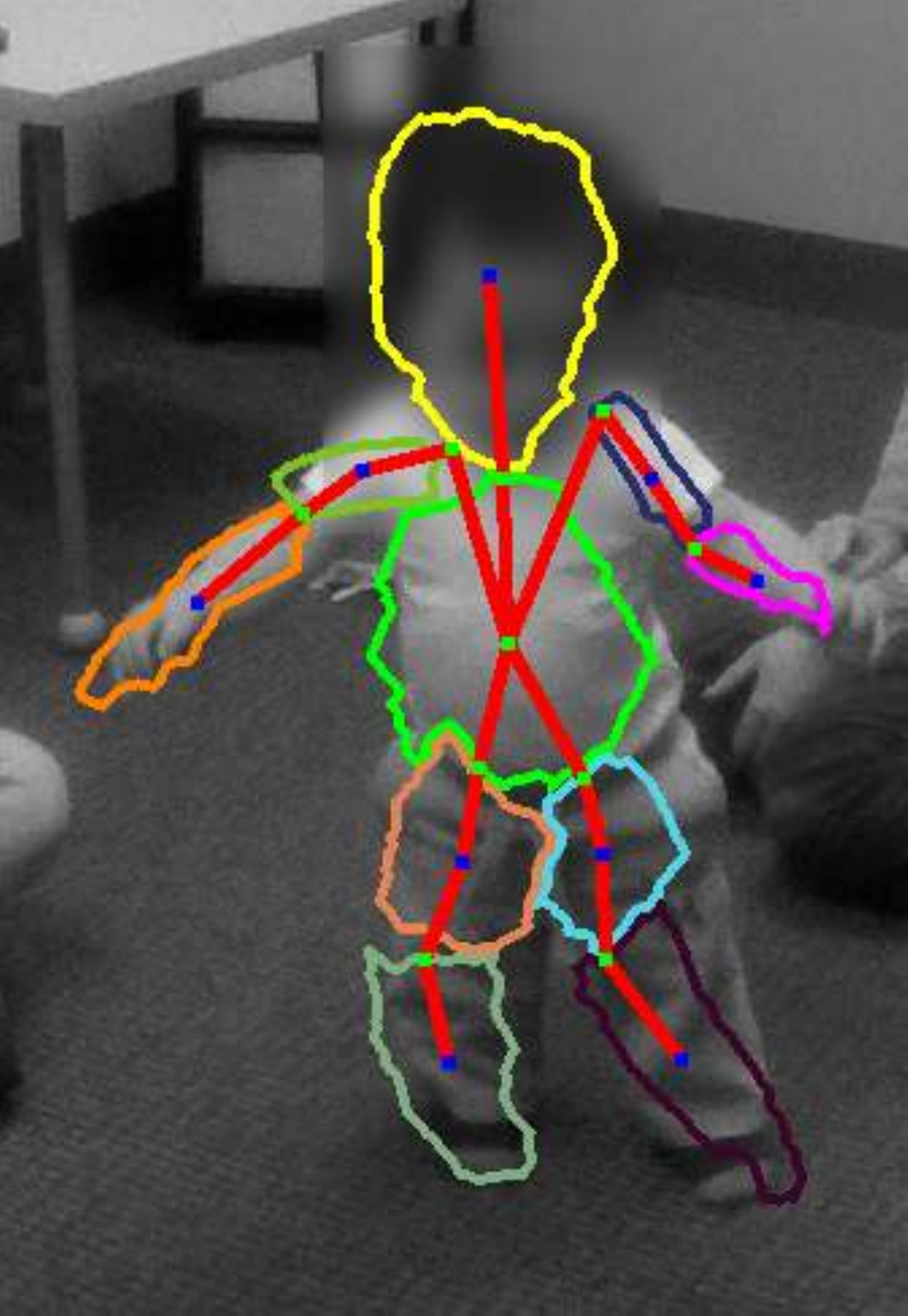}\\ 
  & \fnum{\ti} & \fnum{22} & \fnum{44} & \fnum{88} & \fnum{110} & \fnum{132}\\
  \pthree{} &
  \includegraphics[width=\figresultwidth]{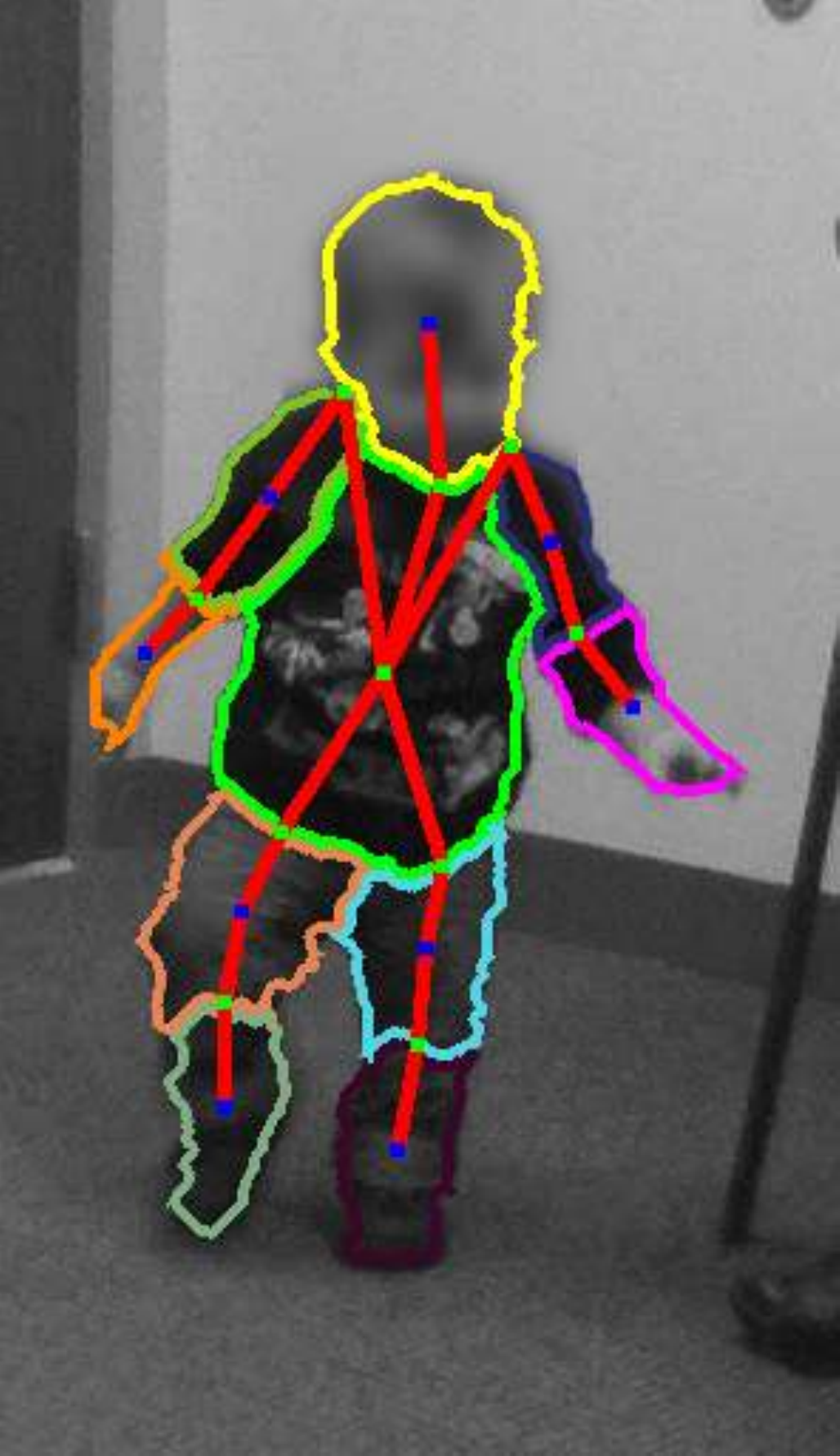} &
  \includegraphics[width=\figresultwidth]{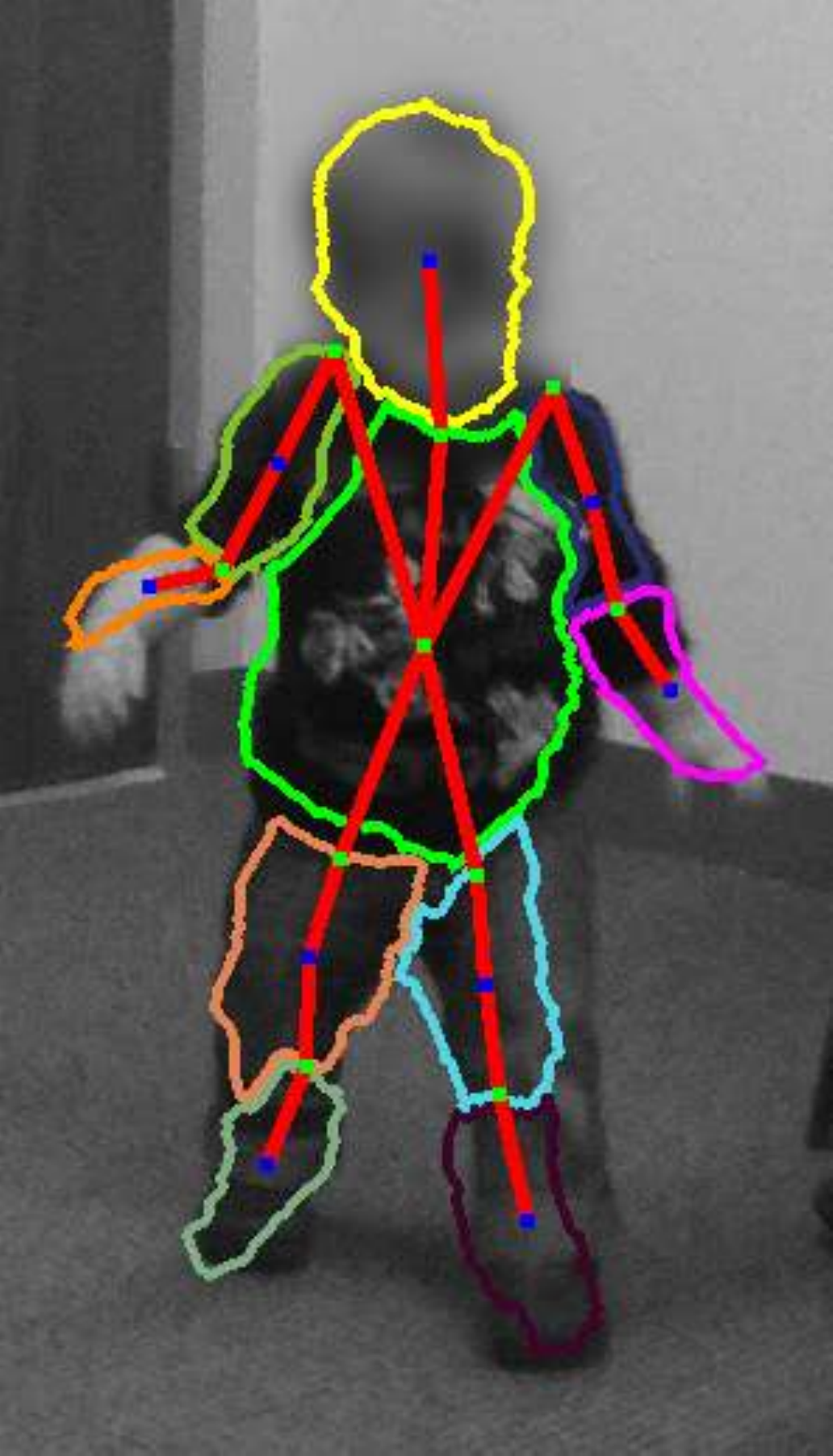} &
  \includegraphics[width=\figresultwidth]{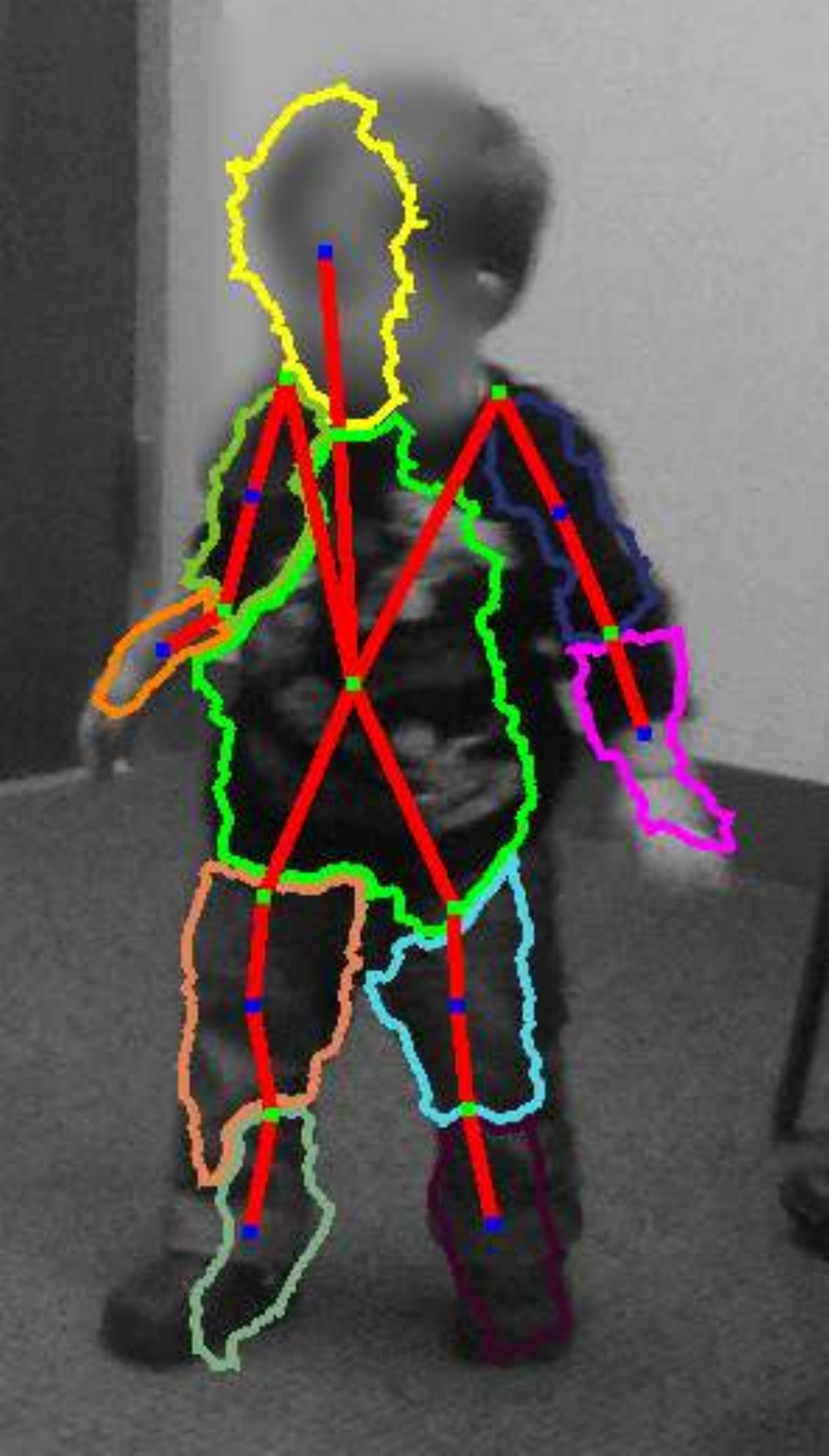} &
  \includegraphics[width=\figresultwidth]{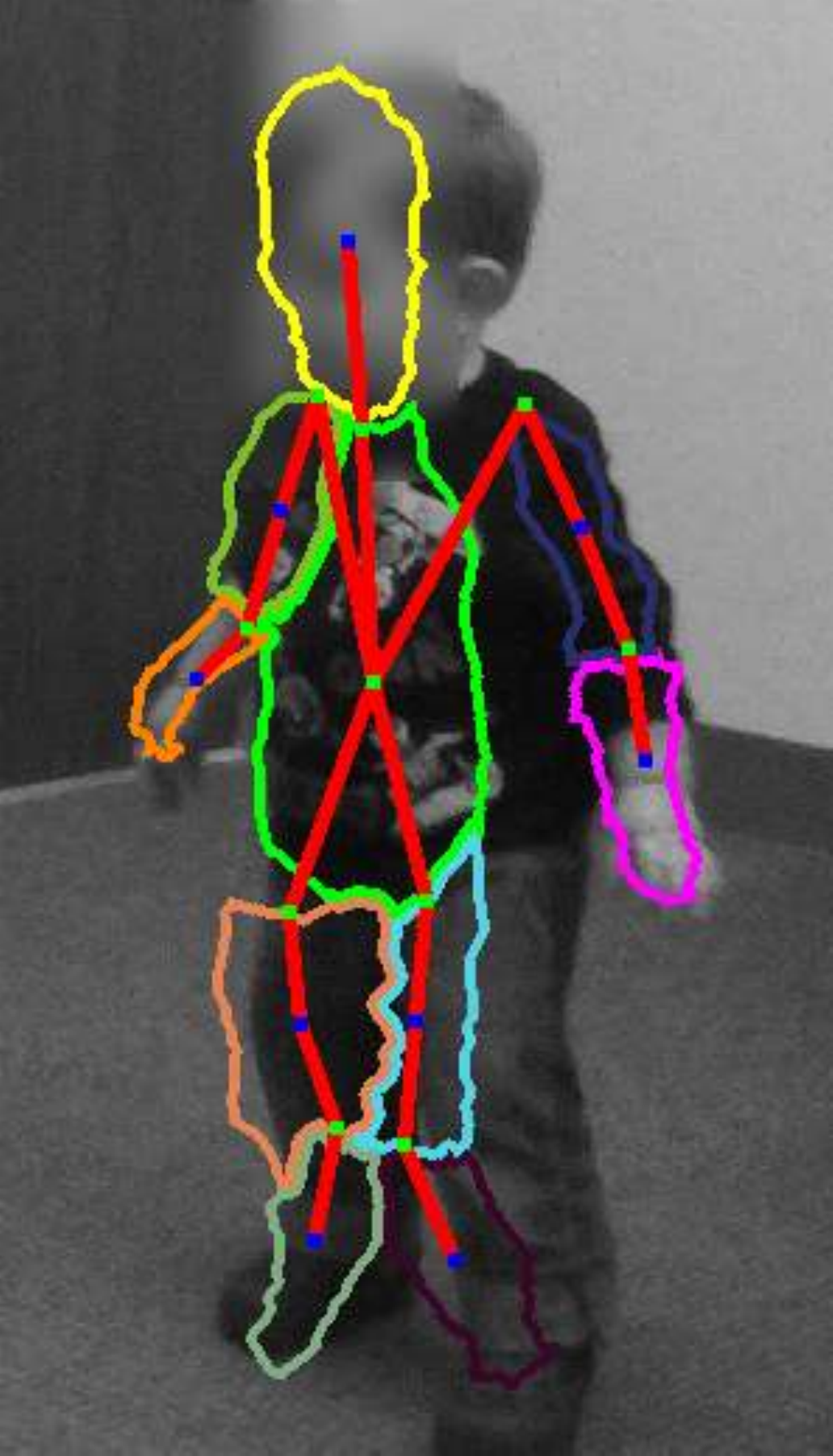} &
  \includegraphics[width=\figresultwidth]{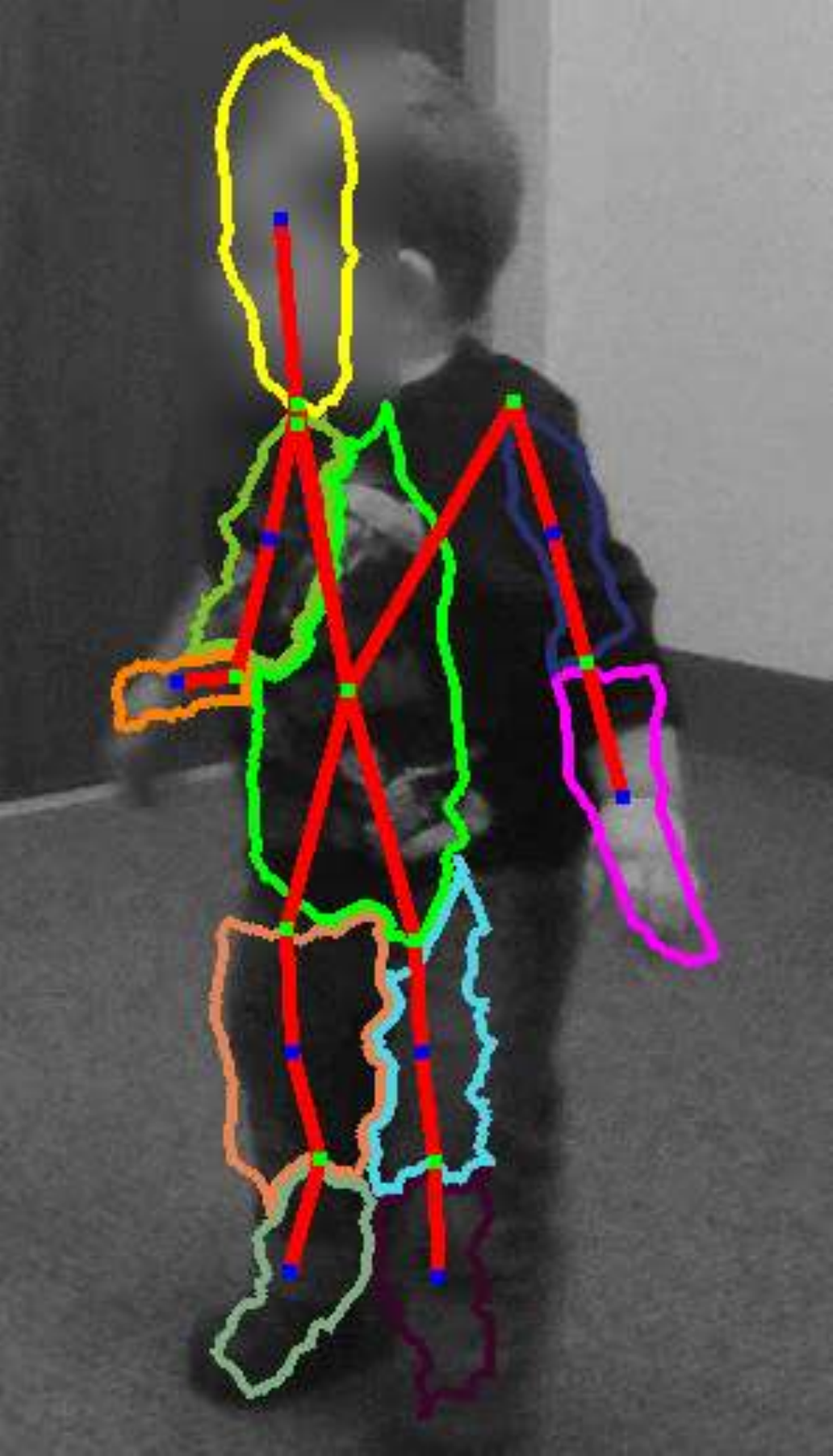} &
  \includegraphics[width=\figresultwidth]{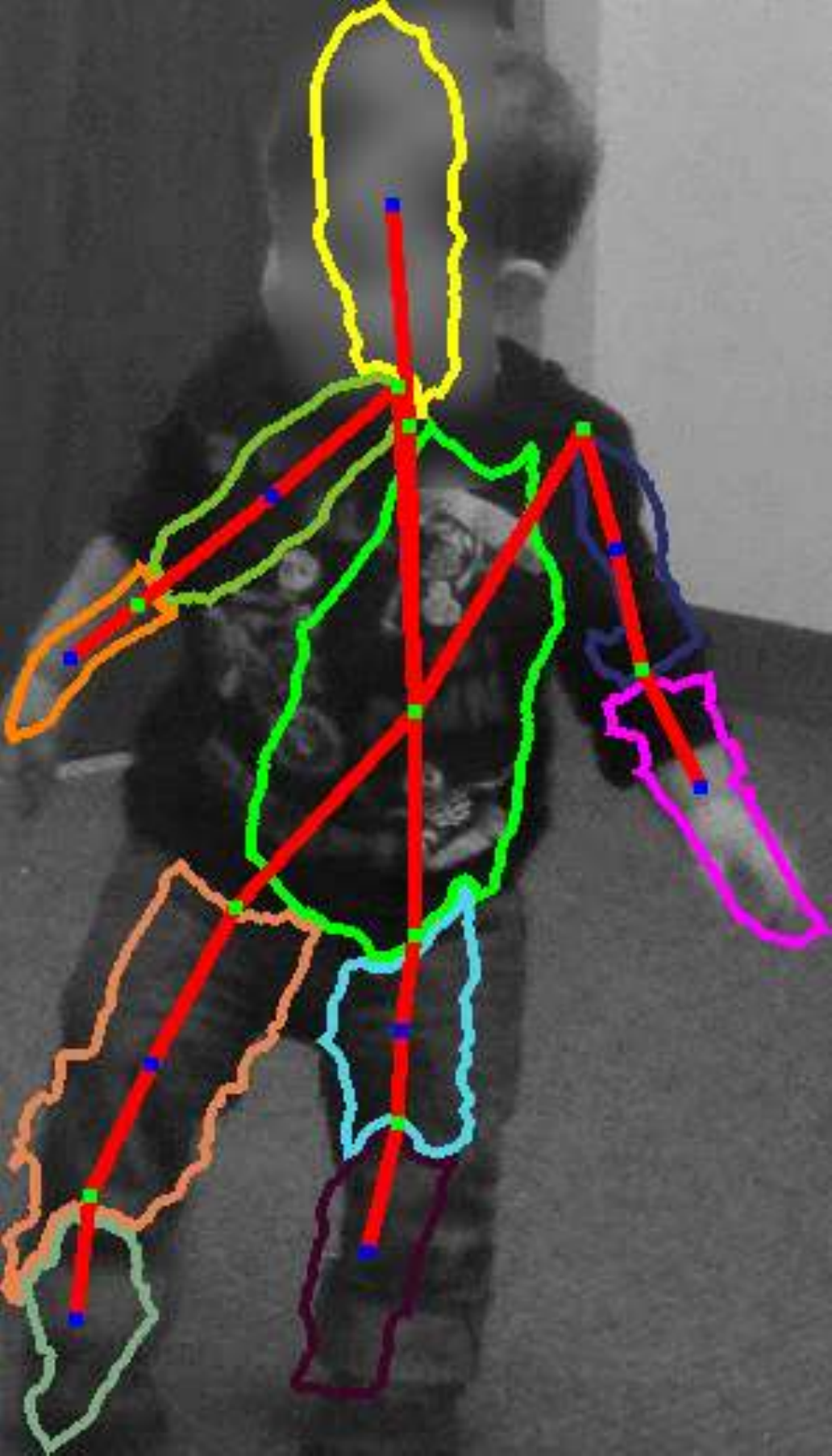} \\ 
  & \fnum{\ti} & \fnum{15} & \fnum{30} & \fnum{45} & \fnum{75} & \fnum{90}\\
  \pone &
  \includegraphics[width=\figresultwidth]{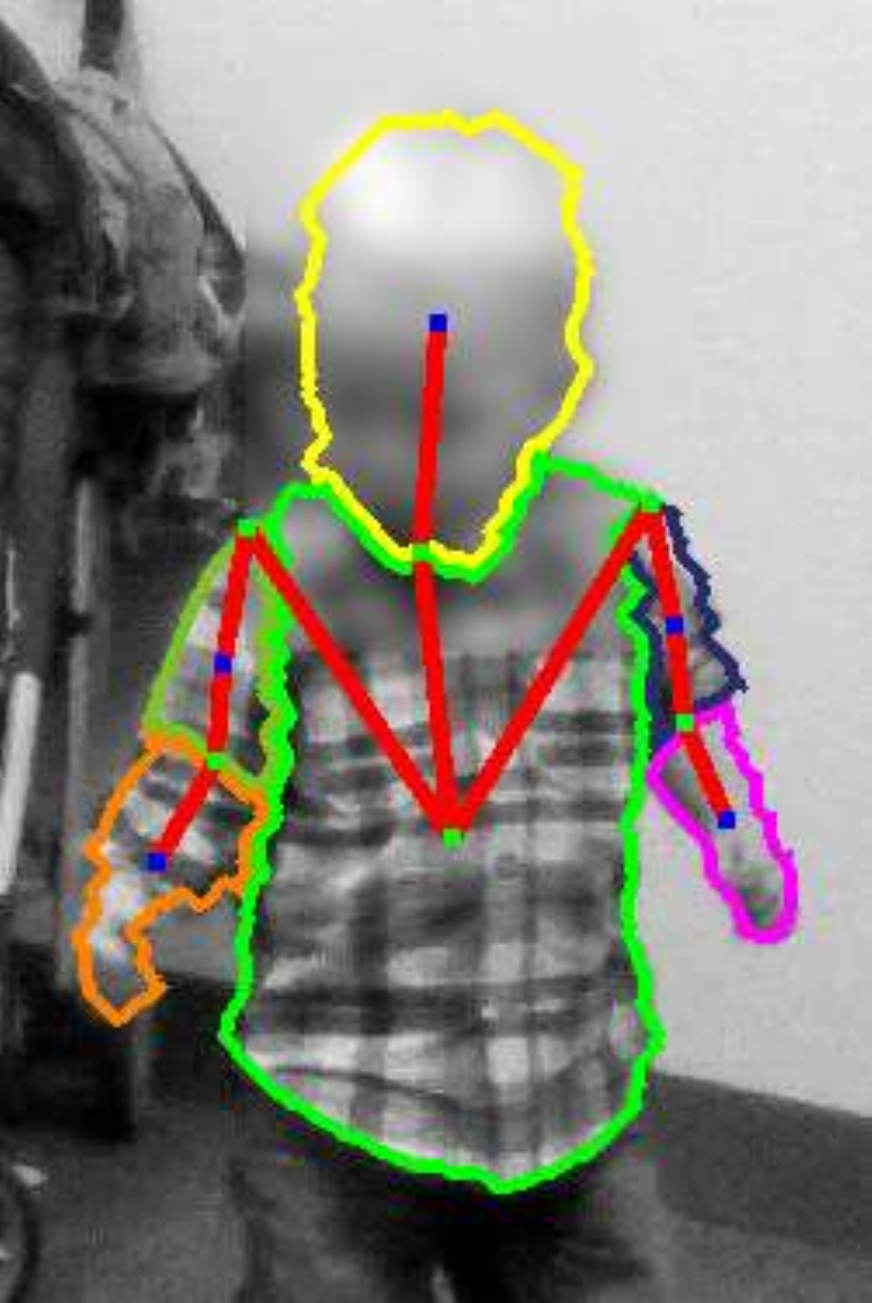} &
  \includegraphics[width=\figresultwidth]{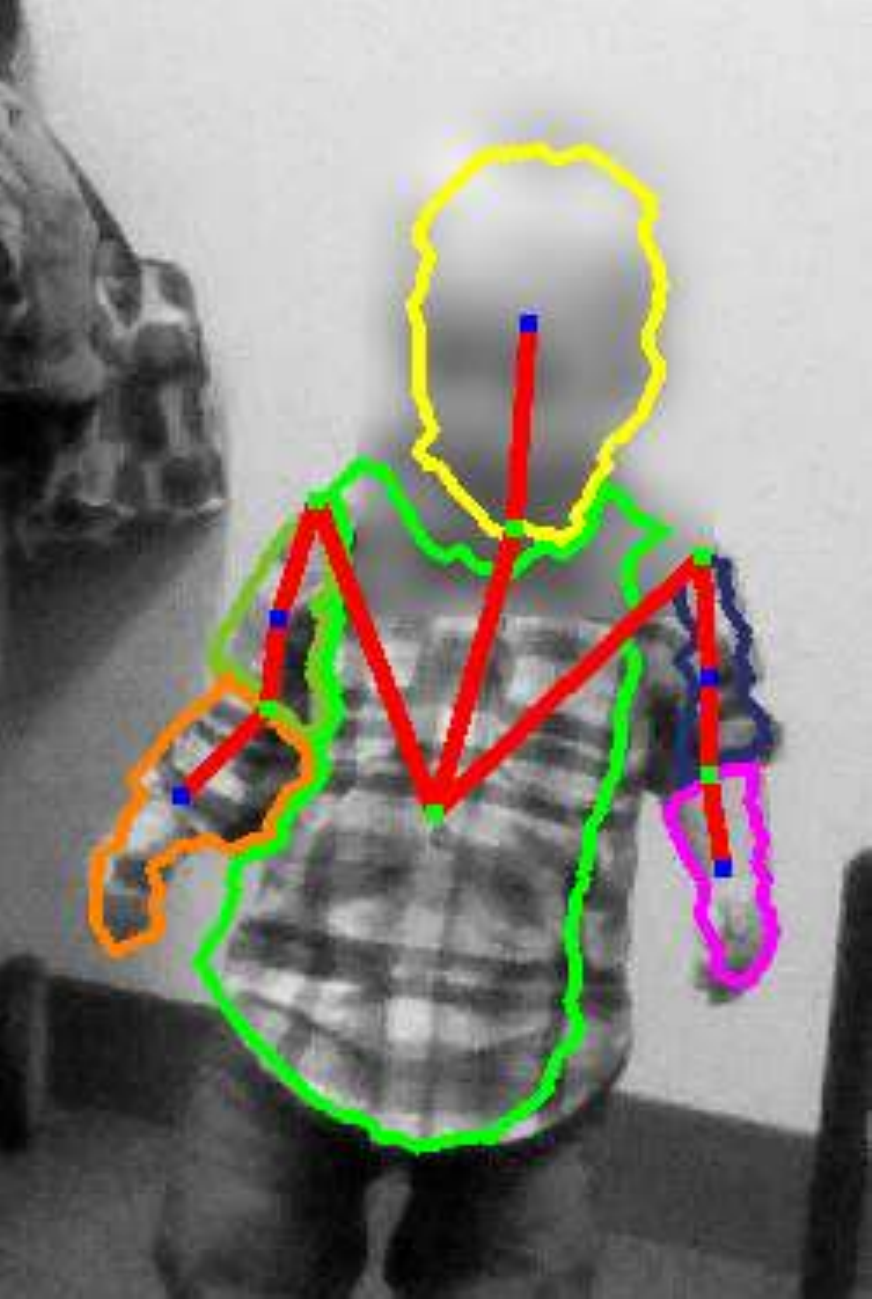} &
  \includegraphics[width=\figresultwidth]{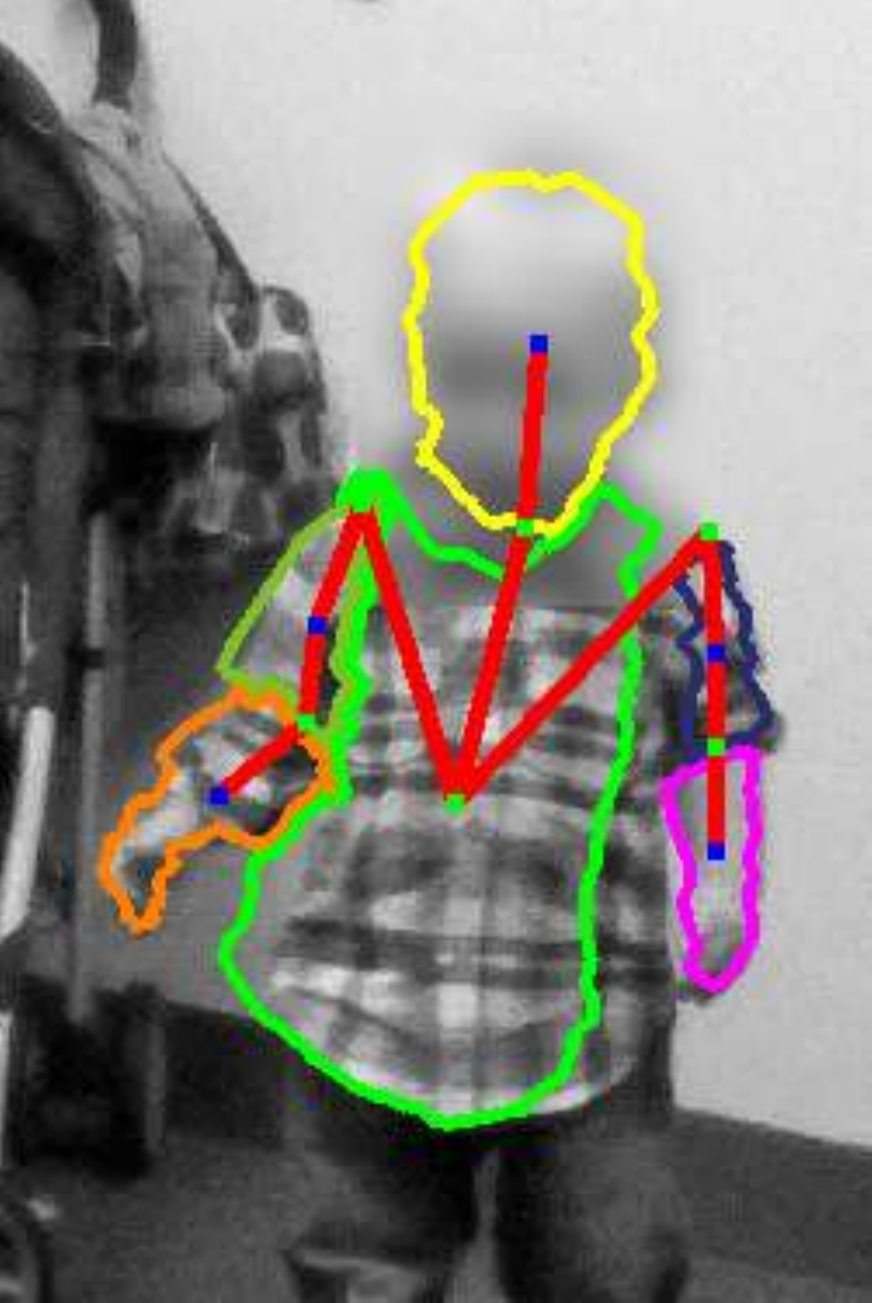} &
  \includegraphics[width=\figresultwidth]{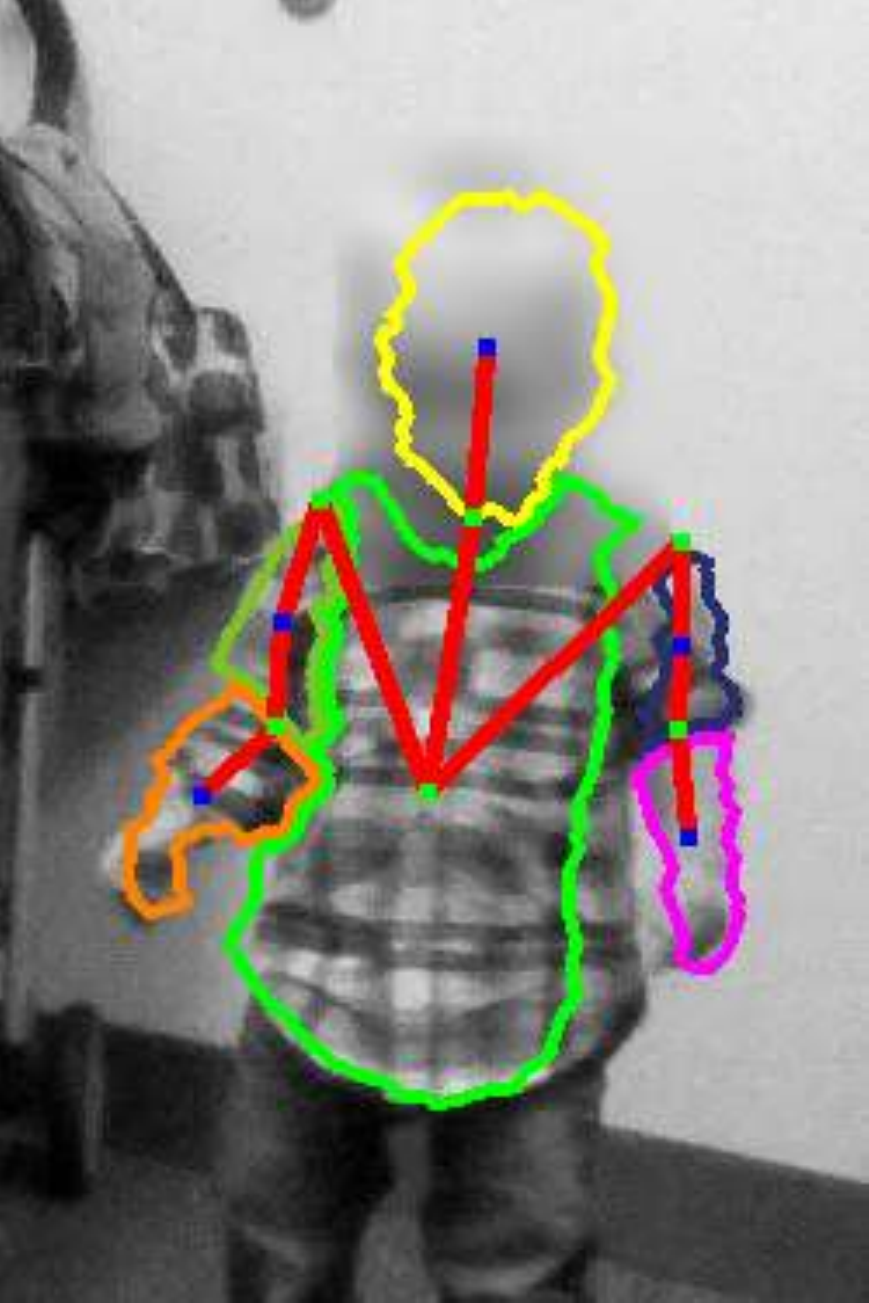} &
  \includegraphics[width=\figresultwidth]{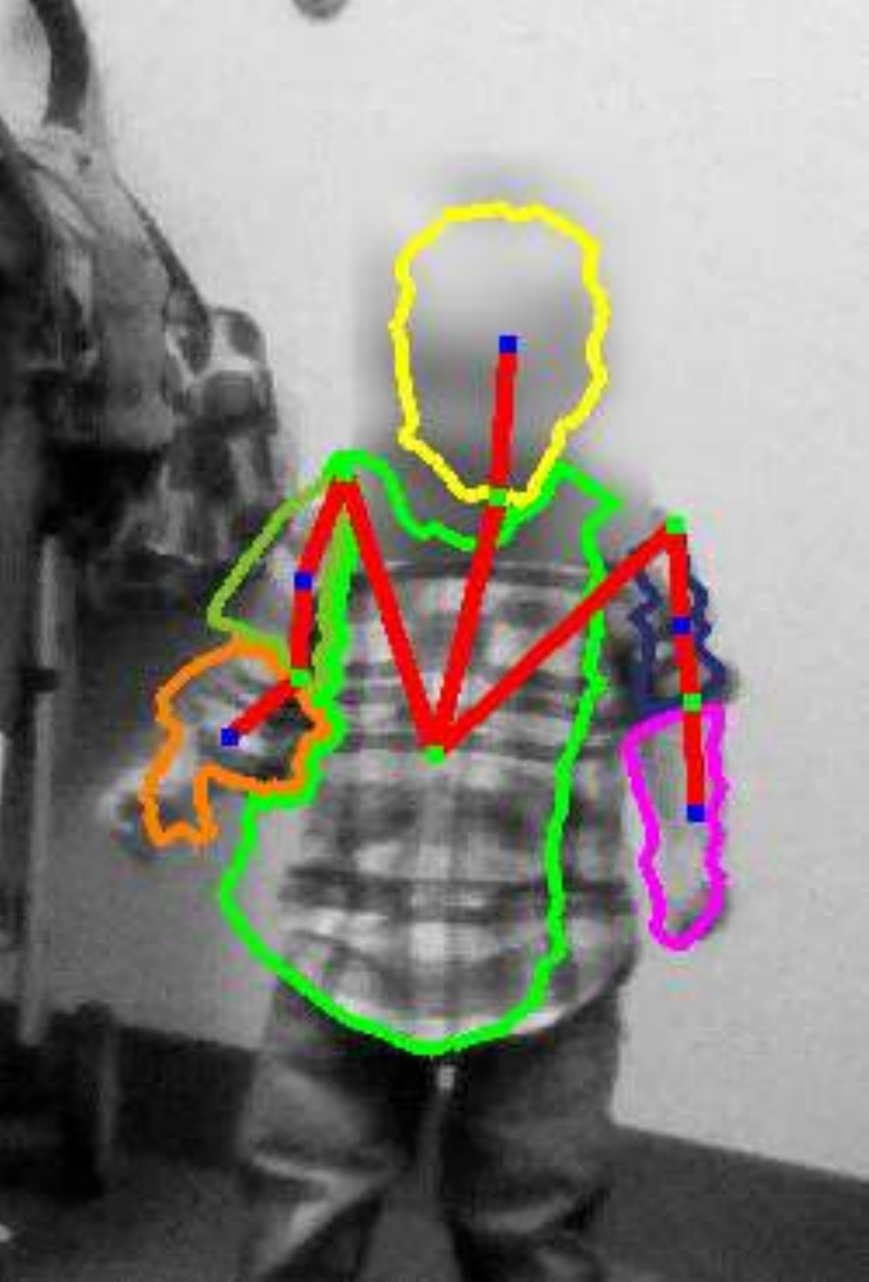} &
  \includegraphics[width=\figresultwidth]{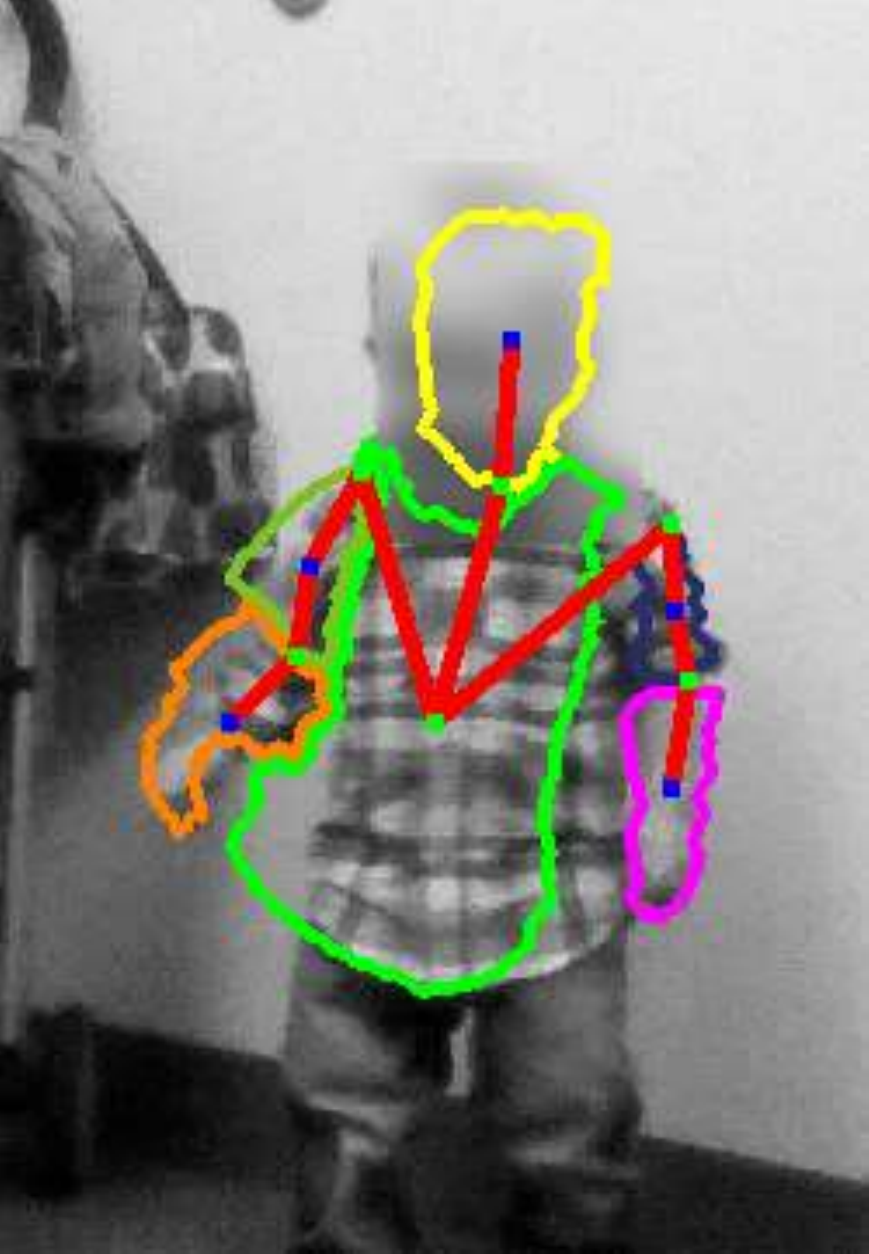}\\
  & \fnum{\ti} & \fnum{24} & \fnum{48} & \fnum{72} & \fnum{96} & \fnum{120} 
\end{tabular}
\end{center}
\caption{Segmentation results using the Cloud System Model (the
  numbers below the images indicate the frames). Even when the
  segmentation starts to fail, latter columns, the estimated stickman
  allows proper body pose estimation.\label{f.segmentation-results}}
\end{figure*}

We tested our human body segmentation algorithm in video clips in
which at least the upper body of the child can be seen, following
Esposito et al.~\cite{Esposito11}
(Figure~\ref{f.segmentation-results}). The result of segmentation is
tightly coupled to the quality of body pose estimation, since the
stickman drives the CSM during the search. However, interactive-level
accuracy is not required from CSM segmentation when performing body
pose estimation for arm symmetry assessment. Hence, our segmentation
algorithm can be comfortably evaluated in such task.

\newcolumntype{C}{>{\centering\arraybackslash}m{\figresultwidth}} 
\newcolumntype{S}{>{\centering\arraybackslash}m{.01\textwidth}} 
\renewcommand{\arraystretch}{0.4}

Our study involves 6 participants, including both males and females
ranging in age from 11 to 16 months.\footnote{Approval for this study
  was obtained from the Institutional Review Board at the University
  of Minnesota. The images displayed here are grayscaled, blurred, and
  downsampled to preserve the anonimity of the
  participants. Processing was done on the original color videos.}  We
have gathered our data from a series of ASD evaluation sessions of an
ongoing concurrent study performed on a group of at-risk infants, at
the Department of Pediatrics of the University of Minnesota. Our setup
includes a GoPro Hero HD color camera positioned by the clinician in a
corner of the room (left image of Figure~\ref{f.pose-estimation}),
filming with a resolution of 1080p at 30 fps. All participants were
classified as a baby sibling of someone with ASD, a premature infant,
or as a participant showing developmental
delays. Table~\ref{part_info} presents a summary of this
information. Note that, the participants are not clinically diagnosed
until they are $36$ months of age and only participant \ptwo{}
(Figure~\ref{f.p02-s01-asymmetry}) has presented conclusive signs of
ASD.
\begin{table}[htb]
\caption{Information on participants involved in this study.}
\centering
\begin{tabular}{cccc}
\toprule

Part \# & Age (months) & Gender & Risk Degree \\ 
\midrule
\pnineteen & 14 & F & Showing delays\\
\pfive & 11 & M & Premature infant\\
\ptwo & 16 & M & ASD diagnosed\\
\pthree{} & 15 & M & Showing delays\\
\pone & 16 & M & Baby sibling\\
\pseven & 12 & F & Premature infant\\
\bottomrule
\end{tabular}
\label{part_info}

\end{table}

We compiled video sequences from ASD evaluation sessions of the 6
toddlers, using one or two video segments to ensure that each child
was represented by one sequence with at least 5s (150 frames). For
each video segment of every sequence, a single segmentation mask was
obtained interactively in the initial frame~\cite{Spina11}. In
contrast, Esposito et al.~\cite{Esposito11} compiled $5$ minutes
sequences at $8$ fps from $50$ participants, that were manually
annotated frame-by-frame using EWMN. Our participants are fewer
(\cite{Esposito11} is a full clinical paper) and our sequences
shorter, though still sufficient, because our dataset does not contain
unsupported gait for longer periods; this is in part because (1) not
all participants evaluated by our clinical expert have reached walking
age and (2) the sessions took place in a small cluttered room (left
image in Figure~\ref{f.pose-estimation}). Hence, we screened our
dataset for video segments that better suited the evaluation of our
symmetry estimation algorithm (with segments of the type used
in~\cite{Esposito11}), rather than considering each child's case. Our
non-optimized single-thread implementation using Python and C++ takes
about 15s per frame (cropped to a size of $\thicksim$500x700px) in a
computer with an Intel Core i7 running at 2.8 GHz and 4GB of RAM.

Table~\ref{t.motor-pattern-eval} summarizes our findings for the $6$
participants. We adopt a strict policy by considering a single frame
asymmetric only when both \ASa{} and \ADf{} agree (i.e., $\ASa\geq1.0$
and $\ADf\geq45^o$) --- see Section~\ref{ss.discussion} for more
information on the adoption of such policy. As aforementioned, we
attempt to quantify asymmetry for each video sequence by computing SS
and DS according to our frame asymmetry policy. 
Table~\ref{t.motor-pattern-eval} also presents the clinician's
visual inspection of each video sequences, categorized as
``symmetric'' (Sym), ``asymmetric'' (Asym), or ``abnormal'' (Abn ---
i.e., some other stereotypical motor behavior is present on the video
segment).
\begin{table*}
\centering
\caption{Symmetry data for the video sequences from $6$ different
  participants used in our experiments. We computed the Static
  Symmetry and Dynamic Symmetry (SS and DS,~\cite{Esposito11}) from
  the automatically obtained skeleton (Aut.), considering a frame
  asymmetric if both \ASa{} and \ADf{} agree (recall that the higher
  the number, the more asymmetrical the walking pattern). We also
  present the Static/Dynamic Symmetry values obtained from the ground
  truth skeleton (GT), the clinician's evaluation about the video
  segments of each sequence, and the video sequence length. For the
  clinician's evaluation, we categorize the results as ``symmetric''
  (Sym), ``asymmetric'' (Asym), or ``abnormal'' (Abn --- i.e., some
  other stereotypical motor behavior is present on the video
  segment). For each video segment, we threshold SS and DS in $30\%$
  to assign a binary grade of asymmetry that can be compared with the
  clinician's assessment (note that SS and DS of individual video
  segments are higher than those of the subsuming video sequences). We
  selected one or two segments for each participant to create
  sequences of at least 5s. \label{t.motor-pattern-eval} }
        {\footnotesize
\begin{tabularx}{\textwidth}{cYYYYYYYYc}
\toprule
\multirow{2}{*}{Part.} & \multicolumn{2}{c}{Stat. Sym. (\%)} & \multicolumn{2}{c}{Dyn. Sym. (\%)} & \multicolumn{2}{c}{Aut. Seq. Eval.} & \multicolumn{2}{c}{Clin. Seq. Eval.} & \multirow{2}{*}{Seq. Length} \\
\cmidrule(rl){2-3} \cmidrule(rl){4-5} \cmidrule(rl){6-7} \cmidrule(rl){8-9}
& Aut. & GT & Aut. & GT & Seg. $1$ & Seg. $2$ & Seg. $1$ & Seg. $2$ & (s.)\\
\midrule
\pnineteen & $36$ & $34$ & $64$ & $55$ & Asym & - & Asym & - & $5.0$\\
\pfive & $0$ & $0$ & $0$ & $0$ & Sym & - & Sym & - & $5.0$ \\
\ptwo & $41$ & $41$ & $44$ & $44$ & Asym & Sym & Asym & Sym/Abn & $7.4$  \\
\pthree{} & $5$ & $0$ & $21$ & $0$ & Sym & Sym & Sym & Sym/Abn & $6.7$ \\
\pone & $0$ & $0$ & $0$ & $0$ & Sym & Sym & Asym & Sym & $7.6$ \\
\pseven & $29$ & $28$ & $36$ & $36$ & Sym & Asym & Sym/Abn & Abn & $6.5$ \\
\bottomrule
\end{tabularx}
}
\end{table*}

\subsection{Discussion}
\label{ss.discussion}

\begin{figure}[htb]
\centering
\subfloat[]{\includegraphics[width=.9\textwidth]{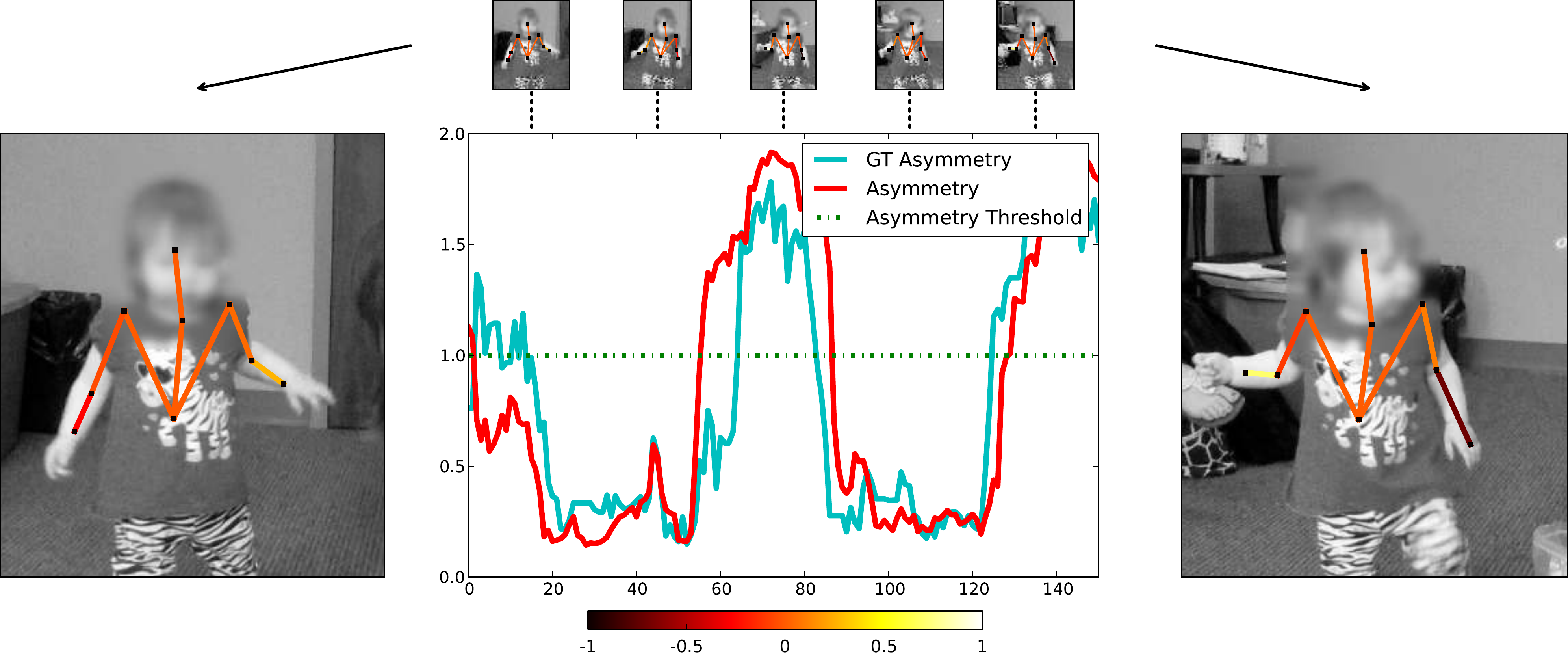} \label{f.p19-s01-asymmetry}}\\ 
\subfloat[]{\includegraphics[width=.45\textwidth]{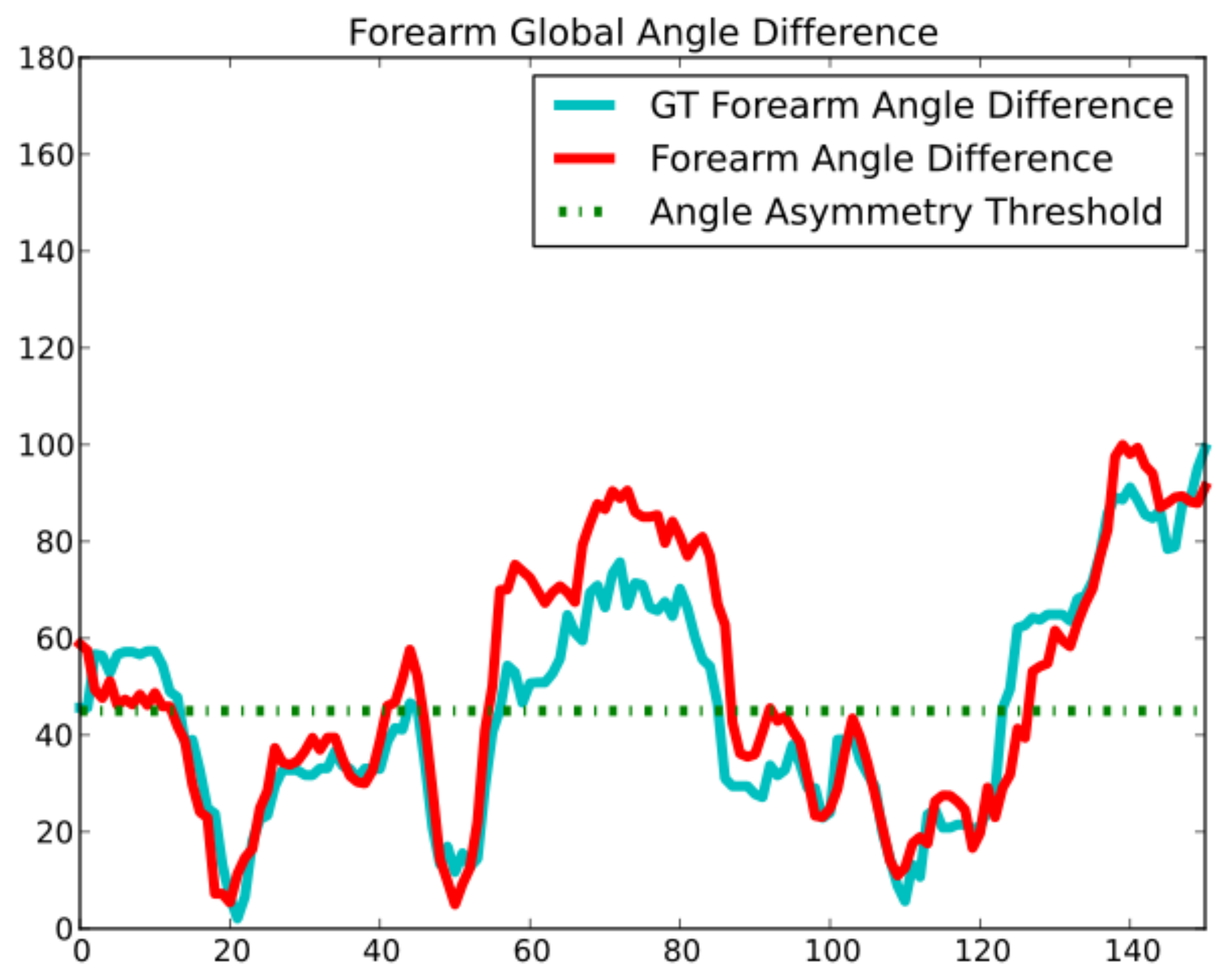} \label{f.p19-s01-angle-diff}}
\caption{(a) Pose estimation performed for a video segment presenting
  participant \pnineteen{} walking unconstrained. We are interested in
  finding when the toddler is walking with asymmetric arm poses, a
  possible sign of ASD. We colorcode the upper arm with the
  corresponding asymmetry score $\ASu$ and the forearm using the final
  asymmetry score \ASa{}, after shifting the mean values to the
  interval $[-1,1]$ to denote the left or right arm segment with
  lowest/highest vertical coordinate. The graph depicts the absolute
  non-shifted final asymmetry score \ASa{} ($y$-axis) across time
  ($x$-axis). We present the asymmetry scores obtained from the ground
  truth skeleton in cyan in the graph. (b) This graph presents the
  difference \ADf{} between the global angle values of participant
  \pnineteen{}'s left and right forearms.\label{f.p19}}
\end{figure}

\begin{figure}[htb]
\centering
\subfloat[]{\includegraphics[width=.9\textwidth]{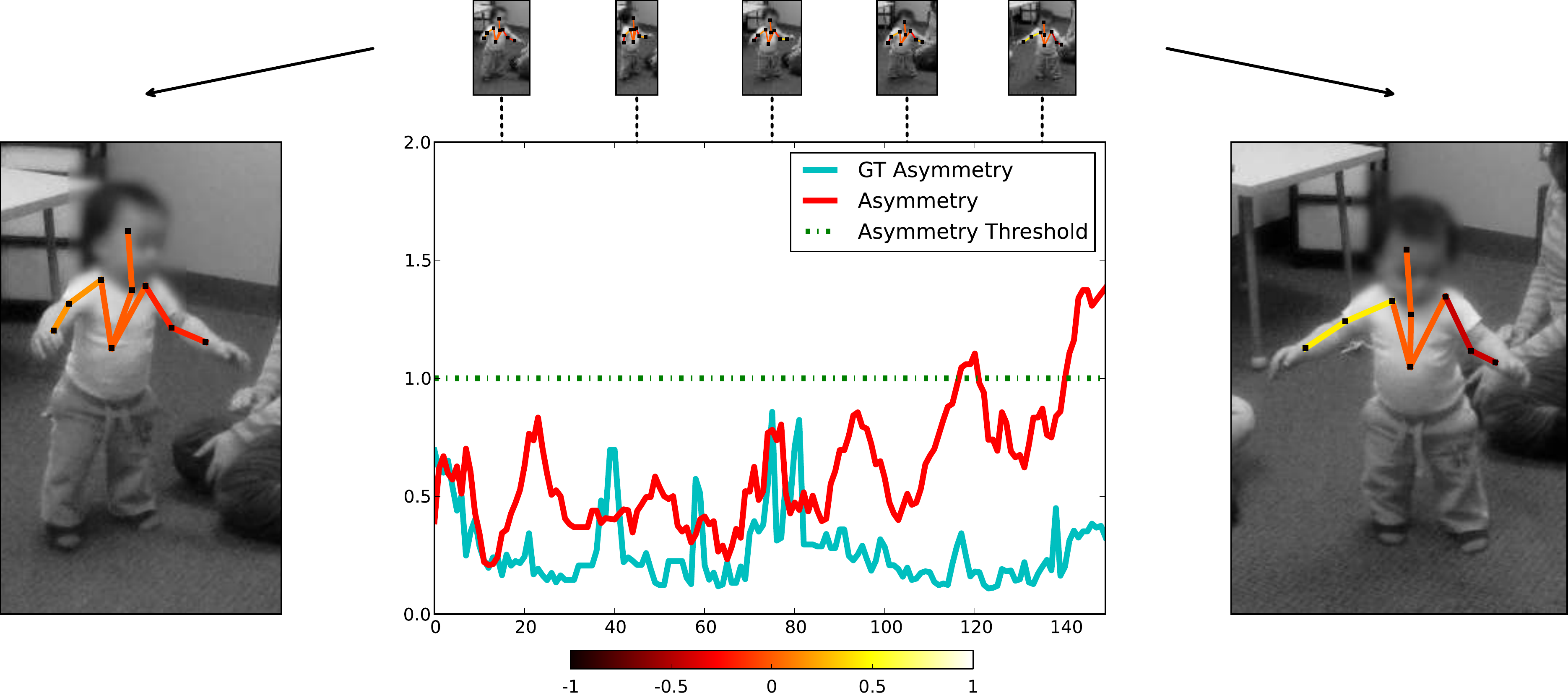} \label{f.p05-s01-asymmetry}} \\
\subfloat[]{\includegraphics[width=.45\textwidth]{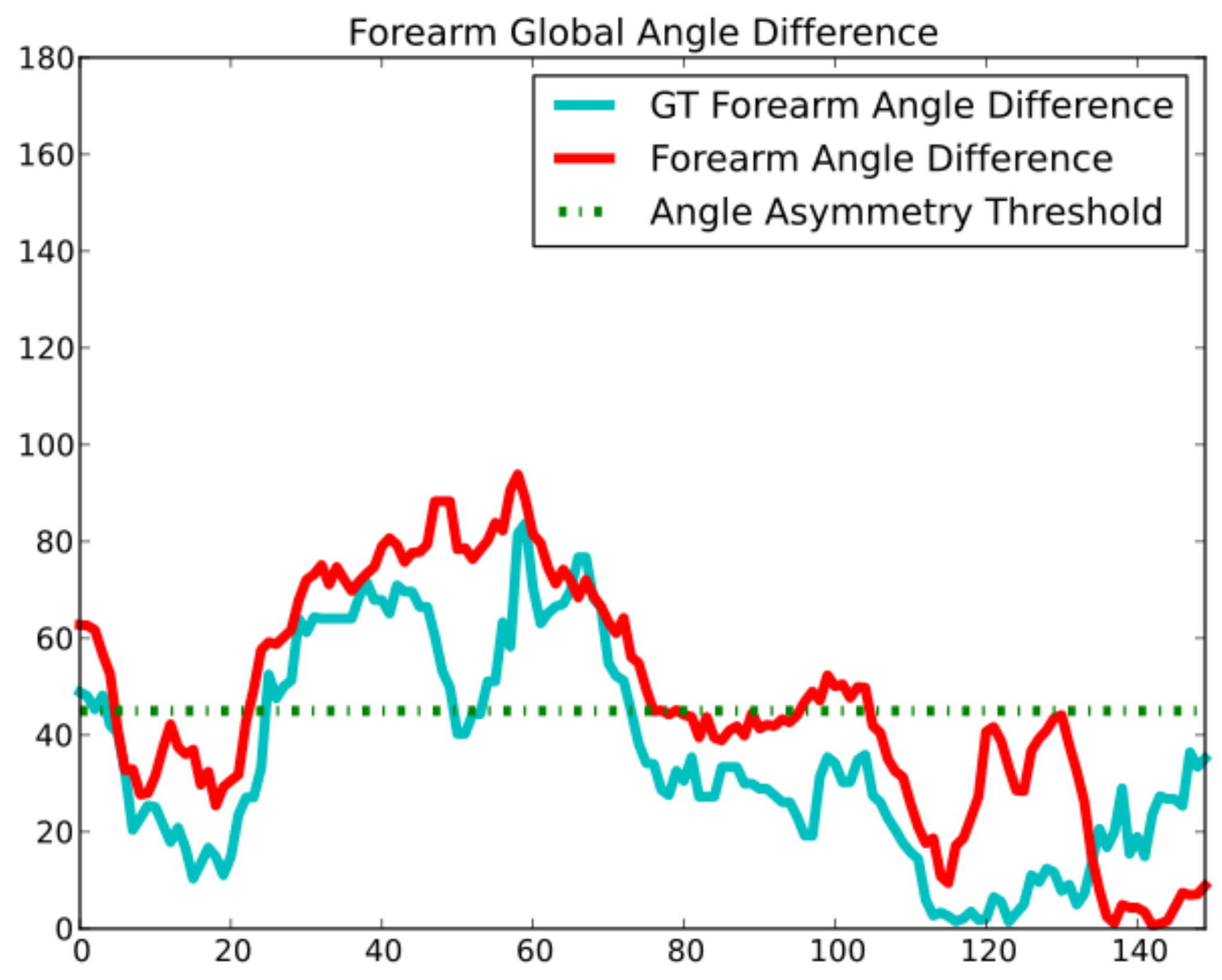} \label{f.p05-s01-angle-diff}}
\caption{(a) \ASa{} asymmetry scores for the video segment from
  participant \pfive. In this video segment, the corresponding \ADf{}
  asymmetry scores presented in (b) indicate false positive asymmetry
  between frames $20-80$, as opposed to the \ASa{}
  scores.\label{f.p05}}
\end{figure}

\begin{figure}
\centering
\subfloat[]{\includegraphics[width=.9\textwidth]{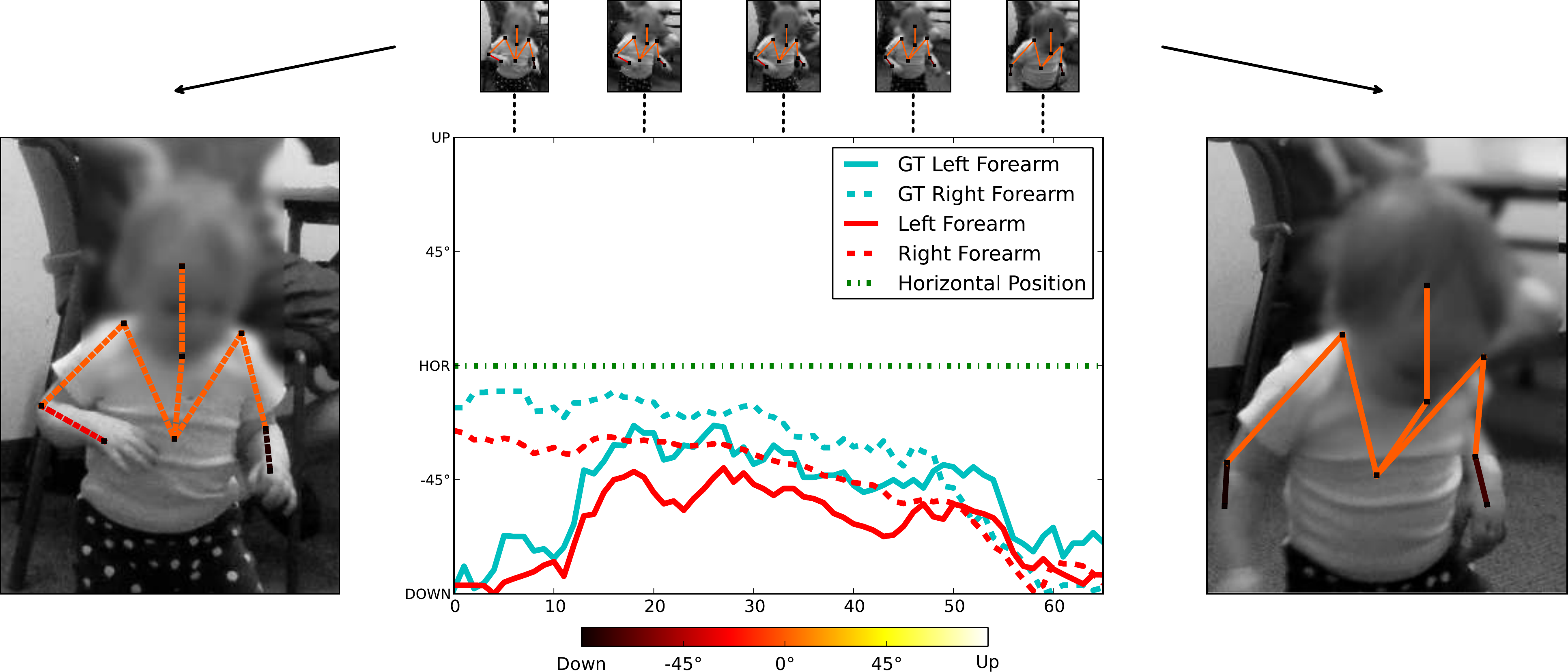} \label{f.p07-s01-angles}}\\ 
\subfloat[]{\includegraphics[width=.9\textwidth]{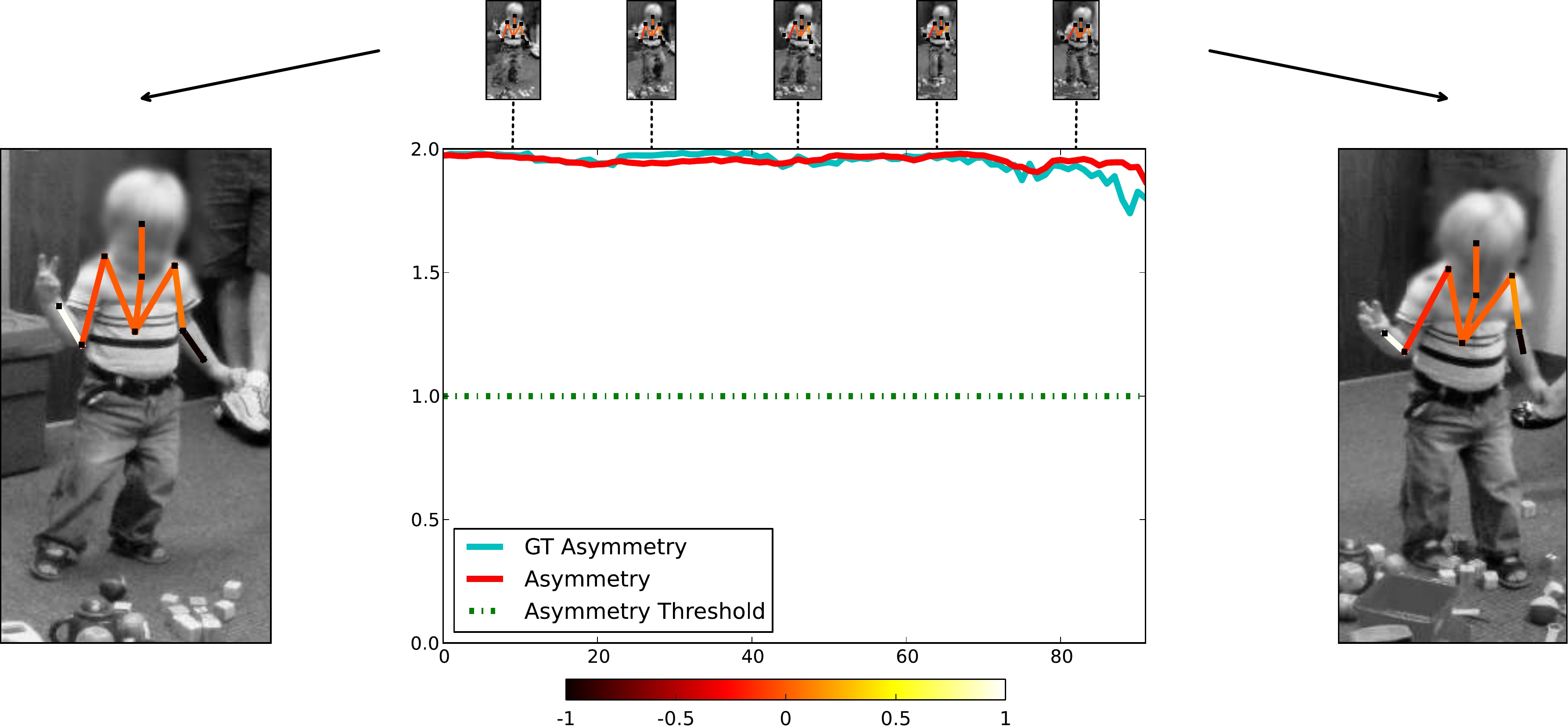} \label{f.p02-s01-asymmetry}}
\caption{(a) First video segment of participant \pseven, where she
  walks holding her arms parallel to the ground pointing forward. The
  graph depicts this behavior by showing the forearm angles w.r.t. the
  horizontal axis. (b) First video segment of participant \ptwo{}, the
  only one diagnosed with autism thus far. In this example,
  participant \ptwo{} is not only presenting asymmetric arm behavior
  throughout the entire video segment, but he is also presenting
  abnormal gait and hand behavior (other types of stereotypical motor
  behaviors).\label{f.p07-p02}}
\end{figure}

Figures~\ref{f.p19}-\ref{f.p07-p02} present
our temporal graphs depicting the asymmetry score \ASa{}, the left and
right forearms' global angles and corresponding difference \ADf{}, as
examples for video segments of 4 participants (with ground truth). The
forearms' global angles essentially denote where each one is pointing
to w.r.t. the horizontal axis (up, down, horizontally).

In Figure~\ref{f.p19-s01-asymmetry}, participant \pnineteen{} walks
asymmetrically holding one forearm in (near) horizontal position
pointing sideways, while extending the other arm downwards alongside
her body in frames $0-18$, $63-85$, and $125-150$. The graph in this
figure represents the asymmetry score \ASa{} computed from both our
automatically computed skeleton (red), and the manually created
ground truth skeleton (cyan). The asymmetry scores from the
automatically computed skeleton and the ones obtained from the
ground truth skeleton correlate for this video segment, demonstrating
the accuracy of the proposed technique. However, since we compute a 2D
skeleton, false positives/negatives might occur due to off-plane
rotations (e.g., the false negative indication of asymmetry between
frames $0$ and $18$). 
Figure~\ref{f.p19-s01-angle-diff} presents the angle difference
measure \ADf{} that might also indicate asymmetry when
$\ADf\geq45^o$~\cite{Hashemi12a}. By analyzing both \ADf{} and \ASa{}
from Figure~\ref{f.p19-s01-asymmetry}, one can often rule out false
positives/negatives that occur (i.e., the aforementioned false
negative indication between frames $0-18$ in
Figure~\ref{f.p19-s01-asymmetry} is captured by the \ADf{} graph in
Figure~\ref{f.p19-s01-angle-diff}).

The example in Figure~\ref{f.p05-s01-asymmetry} of participant
\pfive{} further strenghthens the usage of both \ASa{} and \ADf{} by
depicting a false positive indication of asymmetry. Namely, the
asymmetry scores $AS^*$ between frames $20-80$ denote symmetric
behavior for both the ground truth and our automatically computed
skeleton, while the $\ADf{}\geq60^o$ scores in
Figure~\ref{f.p05-s01-angle-diff} indicate false positive
asymmetry. Such disagreement occurs because \pfive{} walks with his
arms wide open in near frontal view, thereby leading the stickman's
left forearm to appear in horizontal position, while the stickman's
right forearm points vertically down.

Figure~\ref{f.p07-s01-angles} depicts the first video segment of
participant \pseven, in which she walks holding her arms parallel to
the ground pointing forward. The graph depicts this behavior by
showing the forearm angles w.r.t. the horizontal axis. One can notice
the aforementioned stereotypical motor pattern by analyzing from the
graph that both forearms are close to the horizontal position for the
better part of the video. This shows the array of stereotypical
measurements and behaviors we may detect from our body pose estimation
algorithm, of which just a few are exemplified here.

Lastly, in Figure~\ref{f.p02-s01-asymmetry} participant \ptwo{} is not
only presenting asymmetric arm behavior throughout the entire video
segment, but he is also presenting abnormal gait and hand behavior
(other types of stereotypical motor behaviors). We intend to use the
skeleton in the detection of such abnormal behaviors as well, by
extracting different kinds of measures from it.

For all video sequences, our method presents good average correlation
with the ground truth for both \ASa{}, $\bar{r}=0.57$, and \ADf{},
$\bar{r}=0.69$. The correlation of \ASa{} was affected by a negative
score of $r=-0.18$ presented for the first video segment of
participant \pthree{}, which occurred due to oscilations in our
automatically computed skeleton with respect to the ground
truth. Nevertheless, the \ASa{} scores computed for both the skeleton
and the ground truth denoted symmetry for most of the video segment,
agreeing therefore with the clinician's assessment. If we remove the
corresponding video segment, the average \ASa{} increases to
$\bar{r}=0.72$ (and average \ADf{} to $\bar{r}=0.73$), indicating high
correlation. To correlate our results with the clinician's categorical
assessment of each video segment in Table~\ref{t.motor-pattern-eval},
we threshold SS and DS in $30\%$ and deem a video segment asymmetric
when both $SS\geq 30\%$ and $DS\geq 30\%$. We select such value
considering that the average SS for both autistic and non-autistic
children was at least $32\%$ in~\cite{Esposito11}, while the average
DS was at least $26\%$. Our method agrees with the clinician's
categorical assessment in $8$ out of $9$ cases, after excluding video
segment 2 of participant \pseven{} since it is abnormal, with
non-weighted Cohen's kappa inter-rater reliability score of $0.72$
(high).

While our method agrees with the clinician's visual ratings about
symmetry for several cases,
the expert's assessment is based on significantly more data. We
therefore seek and achieve correlation between our results and the
ground truth skeleton to aid in research and diagnosis by
complementing human judgement.
We have further hypothesized that our body pose estimation algorithm
can be used to detect other potentially stereotypical motor behaviors in
the future, such as when the toddler is holding his/her forearms
parallel to the ground pointing forward. Note that the behaviors here
analyzed have only considered simple measures obtained from the
skeleton, whereas we can in the future apply pattern classification
techniques, in particular when big data is obtained, to achieve
greater discriminative power.

\section{Conclusion}
\label{sec:conclusion}

We have developed an extension of the Cloud System Model framework to
do semi-automatic 2D human body segmentation in video. For such
purpose, we have coupled the CSM with a relational model in the form
of a stickman connecting the clouds in the system, to handle the
articulated nature of the human body, whose parameters are optimized
using multi-scale search. As a result, our method performs
simultaneous segmentation and 2D pose estimation of humans in video.

This work is further inserted in a long-term project for the early
observation of children in order to aid in diagnosis of
neurodevelopmental disorders~\cite{Hashemi12a,Hashemi12b,Fasching12}. With the
goal of aiding and augmenting the visual analysis capabilities in
evaluation and developmental monitoring of ASD, we have used our
semi-automatic tool to observe a specific motor behavior from videos
of in-clinic ASD assessment. Namely, the presence of arm asymmetry in
unsupported gait, a possible risk sign of autism. Our tool
significantly reduces the effort to only requiring interactive
initialization in a single frame, being able to automatically estimate
pose and arm asymmetry in the remainder of the video. Our method
achieves high accuracy and presents clinically satisfactory results.

We plan on extending the CSM to incorporate full 3D information using
a richer 3D kinematic human model~\cite{Sherman11}. Of course, there
are additional behavioral red flags of ASD we aim at addressing. An
interesting future direction would be to use our symmetry measurements
to identify real complex motor mannerisms from more typical toddler
movements.\footnote{Bilateral and synchronized arm flapping is common
  in toddlers as they begin to babble, being hard to judge whether
  this is part of normal development or an unusual behavior. This
  issue clearly applies to \pfive's and \pseven's clips from their
  12-month assessments.} This extension also includes detecting ASD
risk in ordinary classroom and home environments, a challenging task
for which the developments here presented are a first step.



\section{Acknowledgments}

We acknowledge Jordan Hashemi from the University of Minnesota, for
his contributions to the clinical aspect of this work. Work supported
by CAPES (BEX 1018/11-6), FAPESP (2011/01434-9 \& 2007/52015-0), CNPq
(303673/2010-9), NSF Grants 1039741 \& 1028076, and the
U.S. Department of Defense.

\bibliographystyle{imsart-number}
\bibliography{intro,csm}

\end{document}